\documentclass[dvipsnames]{article}

    \PassOptionsToPackage{numbers, compress}{natbib}
\usepackage[final]{neurips_2024}
\usepackage{xcolor}

\usepackage{comment}
\usepackage[utf8]{inputenc} % allow utf-8 input
\usepackage[T1]{fontenc}    % use 8-bit T1 fonts
\usepackage{hyperref}       % hyperlinks
\hypersetup{
    colorlinks=true, % AISTATS keeps links in black.
    linkcolor=Blue,
    filecolor=Blue,      
    urlcolor=Blue,
    citecolor=Blue % For biblatex.
    }
\usepackage{url}            % simple URL typesetting
\usepackage{booktabs}       % professional-quality tables
\usepackage{amsfonts}       % blackboard math symbols
\usepackage{nicefrac}       % compact symbols for 1/2, etc.
\usepackage{microtype}      % microtypography
\usepackage{xcolor}         % colors
\usepackage{physics}        % more math commands.
\usepackage{lipsum}         % spacing typesetting.
\usepackage{tikz}
\usetikzlibrary{positioning, shapes.geometric}

\usepackage{amsmath}
\usepackage{amssymb}
\usepackage{mathtools}
\usepackage{amsthm}
\usepackage{mathtools, enumitem}

\usepackage{thm-restate, thmtools}

\newtheorem{theorem}{Theorem}[section]

\newtheorem{definition}[theorem]{Definition}

\newtheorem{condition}[theorem]{Condition}

\usepackage{framed} % highlighted text box
\usepackage{multicol}
\usepackage{graphicx}
\usepackage{subcaption}
\usepackage{appendix}
\usepackage{tikz}
\usetikzlibrary {arrows.meta,arrows,patterns,decorations.pathreplacing,chains,positioning,shapes.geometric,calc} 
\usepackage{pifont}% http://ctan.org/pkg/pifont % for checkmarks, etc

\usepackage[T1]{fontenc} %  enables bold small caps

\usepackage{algorithm}
\usepackage[noend]{algpseudocode}
\usepackage{graphicx}

\usepackage{enumitem}
\setlist[itemize]{noitemsep, topsep=0pt}

\usepackage{changepage}

\AtBeginEnvironment{align}{\setcounter{equation}{0}}
\usepackage{amsmath}
\newcommand{\ind}{\perp\!\!\!\perp} 
\newcommand{\nind}{\not\!\perp\!\!\!\perp}

\usepackage{colortbl} % for colored table columns
\usepackage{array} % for multirows
\usepackage{adjustbox}  % for adjusting width of tables to text width
\usepackage{multirow} % vertical text in table
\usepackage{booktabs} %cmid rule under multicolumn

\usepackage{arydshln} % dashed and dotted hlines
\usepackage{hhline}   % for double cmidrules
\usepackage{setspace} % spacing on table of contents

\title{Hybrid Top-Down Global Causal Discovery \\with Local Search\\ for Linear and Nonlinear Additive Noise Models}

\author{
}

\usepackage[all]{xy}

\author{%
  Sujai Hiremath \\
    Cornell Tech\\%, NY, USA.
  \texttt{sh2583@cornell.edu} \\
  \And
  Jacqueline Maasch \\
  Cornell Tech\\%, NY, USA.
  \texttt{jam887@cornell.edu} \\
  \And
  Mengxiao Gao \\
  Tsinghua University\\%, Beijing, China.
  \texttt{gaomx21@mails.tsinghua.edu.cn} \\
  \And
  Promit Ghosal \\
  University of Chicago\\%, IL, USA.
  \texttt{promit@uchicago.edu} \\
  \And
  Kyra Gan \\
  Cornell Tech\\%, NY, USA.\\
  \texttt{kyragan@cornell.edu} \\
}

\begin{document}

\maketitle
%kyra: use \author{} as illustrated in the template to add authors
%\vspace{-15mm}
% \begin{spacing}{0.5}
% \end{spacing}
\begin{comment}

 \begin{table}[!h]
     \centering
     \begin{tabular}{c c c}
         \textbf{Sujai Hiremath} & \textbf{Jacqueline Maasch} &\textbf{Mengxiao Gao} %JM: removing my middle names to make more room.
         \\
         Cornell University  & Cornell Tech& Tsinghua University
        \\ sh2583@cornell.edu & jam887@cornell.edu & gaomx21@mails.tsinghua.edu.cn
     \end{tabular}
 \end{table}
 \begin{table}[!h]
     \centering
     \begin{tabular}{c c}
     \textbf{Promit Ghosal} & \textbf{Kyra Gan}
         \\
         Brandeis University & Cornell Tech 
         \\ promit@brandeis.edu  & kyragan@cornell.edu
     \end{tabular}
 \end{table}
\end{comment}

\begin{abstract}
    Learning the unique directed acyclic graph corresponding to an unknown causal model is a challenging task. Methods based on functional causal models can identify a unique graph, but either suffer from the curse of dimensionality or impose strong parametric assumptions. To address these challenges, we propose a novel hybrid approach for global causal discovery in observational data that leverages local causal substructures. We first present a topological sorting algorithm that leverages ancestral relationships in linear structural causal models to establish a compact top-down hierarchical ordering, encoding more causal information than linear orderings produced by existing methods. We demonstrate that this approach generalizes to nonlinear settings with arbitrary noise. We then introduce a nonparametric constraint-based algorithm that prunes spurious edges by searching for local conditioning sets, achieving greater accuracy than current methods. We provide theoretical guarantees for correctness and worst-case polynomial time complexities, with empirical validation on synthetic data.%\jm{I might add one sentence at the end here highlighting a key experimental result that makes us stand out, like we had X speedups or X gains in Y metric relative to a baseline.} 
\end{abstract}

%\jm{\textbf{This is my very rough version of the abstract.} Learning the unique directed acyclic graph (DAG) corresponding to an unknown causal structural model is a challenging task. Under the additive noise model, this is often achieved by decomposing structure learning into two phases: topological ordering and edge pruning. However, these approaches often suffer from the curse of dimensionality and high sample complexity, due to high-dimensional regressions and high-order conditional independence testing. To address these challenges, we introduce NAME: a polynomial-time global discovery algorithm for linear non-Gaussian additive noise models. NAME is composed of three phases: root learning, topological ordering, and edge learning. In the first phase, NAME leverages the unique properties of root nodes in additive noise models to identify the roots of the underlying DAG. In the second phase, NAME exploits the root set to infer a hierarchical topological ordering of the causal DAG, which encodes more causal information than prior linear orderings. Lastly, we introduce a nonparametric constraint-based method for edge learning, which offers favorable sample complexity relative to existing methods that rely on sparse regression. We show that NAME is asymptotically guaranteed to return the correct unique DAG for the linear non-Gaussian case. We numerically validate this approach on synthetic data. Results suggest XYZ. Finally, we provide evidence that this approach generalizes to nonlinear settings with arbitrary noise.}
%\vspace{-6pt}
\section{Introduction}
%\vspace{-3pt}
Causal graphical models compactly represent the \emph{data generating processes} (DGP)
of complex systems, including physical, biological, and social domains. Access to the true causal graph or its substructures can offer mechanistic insights \citep{runge_inferring_2019, lee_causal_2022} and enable downstream causal inference, including effect estimation \citep{hoyer_estimation_2008, shah_finding_2022, cheng_toward_2023, gupta_local_2023, maasch2024local, shah2024front}. In practice, the true causal graph is often unknown, and can be challenging to assume using domain knowledge. In such limited-knowledge settings, we can instead rely on causal discovery algorithms that learn the causal graph from observational data in a principled, automated manner \citep{spirtes_causal_2016, glymour_review_2019}.

%Traditional constraint-based approaches to causal discovery infer causal relationships through the conditional independence relations present in the data \citep{spirtes_anytime_2001}. 

%the idea here is to mention the issue with scoring-based methods like GES, otherwise the reviewer might question why we never mentioned them
Traditional approaches to causal discovery infer causal relationships either through conditional independence relations (PC \citep{spirtes_anytime_2001}) or goodness-of-fit measures (GES \citep{chickering_learning_nodate}, GRaSP \citep{grasp_liam_2021}).
While these discovery methods can provide flexibility by not requiring assumptions over the functional form of the DGP, they are generally worst-case exponential in time complexity and learn Markov equivalence classes (MEC) rather than unique \emph{directed acyclic graphs} (DAGs) \citep{montagna_assumption_2023}. Therefore, additional modeling assumptions are often necessary for time-efficient and accurate global discovery. 

% It has been shown that 
Certain parametric assumptions can enable recovery of the unique ground truth DAG, e.g., assuming a particular \emph{functional causal model} (FCM) \citep{zhang_identiability_2009}. Under the \emph{additive noise model} (ANM), we obtain unique identifiability by assuming linear causal mechanisms with non-Gaussian noise distributions \citep{shimizu_linear_2006, shimizu_lingam_2014} or nonlinear causal functions with arbitrary noise \citep{hoyer_nonlinear_2008}. Under the independent additive noise assumption, the causal parents of a variable are statistically independent of its noise term. For this class of models, discovery often entails regressing the variable of interest against its hypothesized parent set and testing for marginal independence between this set and the residual term \citep{peters_causal_2014}. 

Current FCM approaches to global causal discovery trade off between two main issues, suffering from either 1) strong parametric assumptions over the noise or functional form (or both) or 2) the use of high-dimensional nonparametric regressions, which require large sample sizes for reliable estimation and do not scale to large graphs. In addition, current FCM-based methods are ill-suited for causal discovery in sparse causal graphs, a setting that characterizes many high-dimensional applications (e.g., analysis of genetic data in healthcare applications) \citep{gan2021greedy, wong2021decoding, wang2022single, linder2022deciphering}.
%These methods do not take advantage of sparsity in their algorithms, failing to leverage local only constructing a linear topological sort.
%\sujai{something here about global vs local}
%Extensions have also been introduced for pairwise \citep{hyvarinen2013pairwise} and time-series data \citep{hyvarinen2010estimation, kadowaki_estimation_2013}. Interest in Bayesian methods for expressing uncertainty over predicted DAGs \citep{heckerman_bayesian_2006} has given rise to a Bayesian score-based approach \citep{hoyer_bayesian_2009}. Both global \citep{lacerda_discovering_2008, hyttinen_learning_2012}
 %and local \citep{dai2024local} discovery algorithms have sought to relax the acyclicity constraint for learning LiNG SEMs. In the present work, we follow on prior attempts \citep{shimizu2011directlingam} to improve on the efficiency of global LiNG SEM learning in the acyclic and causally sufficient setting.

%kyra: use $\;$ to add space between paragraph title and text to mimic the effect of \paragraph
\textbf{Contributions$\;$}
We propose a hybrid causal discovery approach to graph learning that combines functional causal modeling with constraint-based discovery. We depart from previous methods by characterizing conditions that allow us to search for and exploit local, rather than global, causal relationships between vertices. These local relationships stem from root vertices: this motivates a top-down, rather than bottom-up, approach. Thus, we learn the topological sort and discover true edges starting from the roots rather than the leaves, as in existing methods \citep{peters_causal_2014, rolland_score_2022, montagna_scalable_2023}. This approach leverages sparsity in both the ordering phase and edge discovery phase to reduce the size of conditioning sets, as well as the number of high-dimensional regressions. We summarize our major contributions as follows:
%\vspace{-1mm}
\begin{itemize}[leftmargin=*, itemsep=0pt, parsep=0pt]
\item We introduce a topological ordering algorithm LHTS for linear non-Gaussian ANMs that exploits local ancestor-descendent relationships to obtain a compact hierarchical sort.

%We establish a connection between local ancestral relationships and active causal path relations in data generated by linear non-Gaussian acyclic ANMs, and propose an algorithm that leverages these results to efficiently obtain a compact hierarchical topological ordering.
\item We introduce a topological ordering algorithm NHTS for nonlinear ANMs that exploits local parent-child relationships to run fewer high-dimensional regressions than traditional methods, achieving lower sample complexity.

%We provide evidence that the above approach can generalize to data generated by nonlinear functions with arbitrary noise by considering local parent-child relationships and active causal path relations.
\item We introduce a constraint-based algorithm ED that nonparametrically prunes spurious edges from a discovered topological ordering, leveraging local properties of causal edges to use smaller conditioning sets than traditional sparse regression techniques.

%We establish local properties of causal edges in topological orderings, and propose the first computationally tractable constraint-based method for efficient edge pruning, allowing for the flexibility of incorporating nonparametric tests.
\item We achieve accurate causal discovery in synthetic data, outperforming baseline methods.
\end{itemize}
\textbf{Organization$\;$} 
% First, we discuss related work in below. 
% Section~\ref{sec:related}.
% The rest of the paper is organized as follows. 
After describing the preliminaries in Section \ref{sec:prelim},
% we introduce the basic terms and concepts necessary to understand our work. In Section \ref{sec:lin_setting},
we introduce the linear problem setting in Section \ref{sec:lin_setting}, 
establishing the connection between ancestral relationships and causal active paths and 
introducing a \emph{linear hierarchical topological sorting} algorithm (LHTS).
Next,  we extend our method to the nonlinear setting in Section \ref{sec:nonlin_setting} by
% introduce the nonlinear problem setting, 
establishing the connection between parental relationships and active causal paths, introducing a \emph{nonlinear hierarchical topological sorting} algorithm (NHTS).
In section \ref{sec:edge_discovery}, we establish a sufficient conditioning set for determining edge relations and introduce an efficient \emph{edge discovery} algorithm (ED). We then test LHTS, NHTS and ED in synthetic experiments in Section \ref{sec:empirical}. To conclude, we discuss future work that might generalize our approach to full ANMs.
% \vspace{-6pt}

\textbf{Related Work$\;$}
Our work is related to two kinds of discovery methods that explicitly leverage the topological structure of DAGs: 1) permutation-based approaches, and 2) FCM-based approaches. 

%Recent permutation-based approaches, such as SP \citep{raskutti_2013}, GSP \citep{gsp_solus_2021}, and GRaSP \citep{grasp_liam_2021}, leverage the topological structure of DAGs for efficient causal discovery. 
The original permutation-based approach SP \citep{raskutti_2013} searches over the space of variable orderings to find permutations that induce DAGs with minimal edge counts. Authors in \citep{gsp_solus_2021} introduce greedy variants of SP (such as GSP) that maintain asymptotic consistency; GRaSP \citep{grasp_liam_2021} relaxes the assumptions of prior methods to obtain improvements in accuracy. These methods highlight the importance of using permutations for efficient causal discovery, but generally suffer from the need to bound search runtime with heuristics, poor sample efficiency in high dimensional settings, and are unable to recover a unique topological ordering or DAG (\citep{discovery_review_2024}).

%Additionally, such permutation-based approaches recover only an MEC, potentially containing multiple DAGs with different topological orderings and edges.

On the other hand, the recent stream of FCM-based approaches decompose graph learning into two phases: 1) learning the topological sort, i.e., inferring a causal ordering of the variables; and 2) edge discovery, i.e., 
identifying
edges consistent with the causal ordering
\citep{shimizu2011directlingam, peters_causal_2014, buhlmann_cam_2014, rolland_score_2022, montagna_scalable_2023, sanchez_diffusion_2023, montagna_causal_2023}.

The literature on topological ordering algorithms for ANMs is organized along the types of parametric assumptions made on both the functional forms and noise distributions of the underlying DGP.
Early approaches like ICA-LiNGAM \citep{shimizu_linear_2006} and DirectLiNGAM \citep{shimizu2011directlingam} focus on learning DAGs generated by linear functions and non-Gaussian noise terms. 
% A recent body of
Recent work leverages score matching to obtain the causal ordering in settings with nonlinear functions and Gaussian noise: SCORE \citep{rolland_score_2022} and DAS \citep{montagna_scalable_2023} exploit particular variance properties, while DiffAN estimates the score function with a diffusion model \citep{sanchez_diffusion_2023}. NoGAM \citep{montagna_causal_2023} generalizes the score-matching procedure of SCORE
to
% designed for data generated by
nonlinear causal mechanisms 
with
% and
arbitrary noise distributions. RESIT \citep{peters_causal_2014} leverages residual independence results in nonlinear ANMs to identify topological orderings when the noise distribution is arbitrary.
NoGAM and RESIT both rely on high-dimensional nonparametric regression. 

Once a topological ordering is obtained,
spurious edges are pruned. 
Works that are agnostic to the distribution of noise often use a parametric approach, implementing either a form of sparse regression (e.g., Lasso regression \citep{lasso_tibshirani}) or a version of additive hypothesis testing with
% by
generalized additive models (GAMs) \citep{additive_hypothesis_marra} (e.g., CAM-pruning \citep{buhlmann_cam_2014}). RESIT \citep{peters_causal_2014} provides another alternative edge pruning procedure for nonlinear ANMs, relying again on high-dimensional nonparametric regression.
%To our knowledge, RESIT \citep{peters_causal_2014} is the only algorithm that provides nonparametric edge discovery,relying again on high-dimensional nonparametric regression.

%This work introduces a global causal discovery method that takes a hybrid approach to graph learning by combining functional causal modeling with constraint-based discovery. This method is biphasic: Phase 1 discovers all pairwise ancestral relations under the linear non-Gaussian acyclic model (LiNGAM) using marginal independence tests and (un)conditional pairwise regressions, returning a compact hierarchical topological sort; Phase 2 learns the directed edge set using conditional independence testing. Relative to existing methods that combine topological ordering with edge pruning, our method minimizes the cardinality of the hypothesized ancestor sets used in both Phase 1 and 2. Thus, the design of this method mitigates the impacts of sample complexity and the curse of dimensionality on error propagation in estimation, resulting in a performant discovery algorithm for high-dimensional sparse graphs in the finite sample regime.

%\vspace{-5pt}
\section{Preliminaries}
\label{sec:prelim}
%\vspace{-3pt}
We focus on \emph{structural causal model}s (SCMs) represented as DAGs. These graphs describe the causal relationships between variables, where an edge $x_i \rightarrow x_j$ implies that $x_i$ has a direct causal influence on $x_j$.
Let  \( G = (V, E) \) be a DAG on $|V| = d$ vertices, where \( E \) represents directed edges. To define pairwise relationships between vertices, we let 
\(\text{Ch}(x_i)\) denote the children of $x_i$ such that $x_j \in \text{Ch}(x_i)$ if and only if $ x_i \to x_j$, and \(\text{Pa}(x_i)\) denote the parents of $x_i$ such that $x_j \in \text{Pa}(x_i)$ if and only if $x_j \to x_i$.
Similarly, let \(\text{An}(x_i)\) denote the ancestors of $x_i$ such that \(x_j \in \text{An}(x_i)\) if and only if there exists a directed path $x_j \dashrightarrow x_i$, and \(\text{De}(x_i)\) denote the descendants of $x_i$ such that \(x_j \in \text{De}(x_i)\) if and only if there exists a directed path $x_i \dashrightarrow x_j$. 
Vertices can be classified based on the totality of their pairwise relationships: $x_i$ is a \textit{root} if and only if \(\text{Pa}(x_i)=\emptyset\), a \textit{leaf} if and only if \(\text{Ch}(x_i)=\emptyset\), an isolated vertex if $x_i$ is both a root and a leaf, and an intermediate vertex otherwise. See an illustrative DAG in Figure~\ref{fig:example_DAG}.
Vertices can also be classified in terms of triadic relationships: $x_i$ is a confounder of $x_j, x_k$ if and only if $x_i \in \text{An}(x_j)\cap \text{An}(x_k)$; a mediator of $x_j$ to $x_k$ if and only if $x_i \in \text{De}(x_j)\cap \text{An}(x_k)$; and a collider between $x_j$ and $x_k$ if and only if $x_i \in \text{De}(x_j)\cap \text{De}(x_k)$. 
% We provide an 

Undirected paths that transmit causal information between vertices $x_j,x_k$ can be differentiated into \textit{frontdoor} and \textit{backdoor paths} \citep{peter_spirtes_causation_2000}. A frontdoor path is a directed path $x_j \dashrightarrow \cdots \dashrightarrow x_k$ that starts with an edge out of $x_j$, and ends with an edge into $x_k$. A backdoor path is a path $x_j\dashleftarrow \cdots \dashrightarrow x_k $ that starts with an edge into $x_j$, and ends with an edge into $x_k$. Paths that start and end with an edge out of $x_j$ and $x_k$ ($x_k\dashrightarrow \cdots \dashleftarrow x_k $) do not transmit causal information between $x_j,x_k$.

Paths between two vertices are further classified, relative to a vertex set \(\textbf{Z}\), as either \textit{active} or \textit{inactive} \citep{peter_spirtes_causation_2000}. A path between vertices $x_j, x_k$ is active relative to \(\textbf{Z}\) if every node on the path is active relative to \(\textbf{Z}\). Vertex $x_i$ on a path is active if one of the following holds: 1) $x_i$ is not a collider and $x_i \not \in \textbf{Z}$, 2) $x_i$ is a collider and $x_i \in \textbf{Z}$, 3) $x_i$ is a collider and $x_i \not \in \textbf{Z}$, but $\text{De}(x_i) \cap \textbf{Z} \neq \emptyset$.
%1) $x_i \in \textbf{Z}$ and $x_i$ is a collider, 
%or 2) $x_i \not \in \textbf{Z}$ and $x_i$ is not a collider, 
%or 3) $x_i \not \in \textbf{Z}$, $x_i$ is a collider, but $\text{De}(x_i) \cap \textbf{Z} \neq \emptyset$. 
An inactive path is simply a path that is not active. Following convention, throughout the rest of the paper
we will describe causal paths as active or inactive with respect to $\textbf{Z} = \emptyset $ unless 
otherwise specified.

\begin{definition}[Topological Orderings]\label{def:topo_order}
Consider a given DAG $G = (V,E)$. A topological sort (linear order) is a mapping $\pi: V \rightarrow \{0,1,\ldots, |V|-1\}$, 
such that if $x_i \in \text{Pa}(x_j)$, then $x_i$ appears before $x_j$ in the sort $\pi$: $\pi(x_i) < \pi(x_j)$.
A hierarchical sort (between a partial and linear order) 
is  a mapping
$\pi_L: V \rightarrow \{0,1,\ldots, |V|-1\}$, 
such that if $\text{Pa}(x_i)=\emptyset$, then $\pi_L(x_i)=0$, and if $\text{Pa}(x_i)\neq \emptyset$, then $\pi_L(x_i)$ equals the maximum length of the longest directed path from each root vertex to $x_i$, i.e., $\pi_L(x_i)= 1 + \max{\{\pi_L(x_j): x_j \in \text{Pa}(x_i)\}}$. 
\end{definition}

We note that the hierarchical sort is unique, and that it coincides with a topological sort
when the number of layers equals $|V|$, i.e., the DAG is complete.

\begin{definition}[ANMs]\label{def:anm}
ANMs \citep{hoyer_bayesian_2009} are a popular general class of SCMs defined over a DAG $G$ with
% where
%\vspace{-5pt}
\begin{align}\label{eq:ANM}
    x_i = f_i(\text{Pa}(x_i)) + \varepsilon_i, \forall x_i \in V,
\end{align}
where $f_i$s are arbitrary functions and $\varepsilon_i$s are independent arbitrary noise distributions.
\end{definition}
 This model implicitly assumes the causal Markov condition and acyclicity; %and that all variables $x_i \in V$ are observed; %for any observed variable $x_i \in V$, there does not exist any unobserved cause of $x_i$ 
  we adopt the aforementioned assumptions, as well as faithfulness \citep{spirtes_causal_2016}.
%%%%%%%%%%%%%%%%%%%%%%%%

\begin{figure}[t]
\centering
\begin{tikzpicture}[scale=0.1, every node/.style={inner sep=0pt}]
% Nodes
\draw [magenta, fill=magenta, fill opacity=0.2] (0,0) circle (3); % Z_1
\draw (0,0) node {$x_1$};
\draw [WildStrawberry, fill=WildStrawberry, fill opacity=0.2] (15,0) circle (3); % X
\draw (15,0) node {$x_2$};

\draw [pink, fill=pink, fill opacity=0.2] (30,0) circle (3); % Y
\draw (30,0) node {$x_3$};

% Arrows
% Z_1 to X
\draw [black, -{Latex[length=2mm,width=4pt]}] (3,0) -- (12,0);

% Z_1 to Y (curving around X)
\draw [black, -{Latex[length=2mm,width=4pt]}] (3,0) to[bend left=45] (27,0);

% X to Y
\draw [black, -{Latex[length=2mm,width=4pt]}] (18,0) -- (27,0);
% % Label below (optional)
% \node at (15, -8) {\footnotesize \textbf{Figure 1:} Exemplary DAG,  where \(x_1\) is a root, \(x_3\) is a leaf, \(x_3 \in \text{Ch}(x_2), x_3 \in \text{De}(x_1), \)};
\end{tikzpicture}
\caption{Illustrative 
DAG,  where \(x_1\) is a root, \(x_3\) is a leaf, \(x_3 \in \text{Ch}(x_2), x_3 \in \text{De}(x_1). \)}
\label{fig:example_DAG}
%\vspace{-10pt}
\end{figure}
%\vspace{-5pt}
\section{Linear Setting}\label{sec:lin_setting}
%\vspace{-3pt}
We first restrict our attention to ANMs that feature only linear causal functions $f$, known as Linear Non-Gaussian Acyclic causal Models (LiNGAMs).
Following \citep{shimizu2011directlingam}, we note that a LiNGAM can be represented as a $d \times d$ adjacency matrix $B = \{b_{ij}\}$, where $b_{ij}$ is the coefficient from $x_j$ to $x_i$. Note that, for any topological ordering $\pi$ of a LiNGAM, if $\pi(x_j)>\pi(x_i)$, then $b_{ji} =0$. 
Thus, each $x_i \in V$ admits the following compact representation: $x_i = \sum_{\pi(x_j)<\pi(x_i)}b_{ij}x_j + \varepsilon_i.$

\textbf{Identifiability$\;$}
Identifiability conditions for LiNGAMs \citep{shimizu_linear_2006} primarily concern the distribution of errors $\varepsilon_i$: under Gaussianity, distinct linear DGPs can admit the same joint distribution, making them impossible to distinguish. \citet{shimizu_linear_2006} generalize this intuition with \emph{independent component analysis} (ICA) \citep{comon_ica_1994} to provide a multivariate identifiability condition for LiNGAMs (see Appendix \ref{appendix:lin_identify}). In this section, we adopt the aforementioned condition.
%%%%%%%%%%%%%%%%%

%%%%%%%%%%%%%%%%%
\textbf{Ancestral Relations and Active Causal Paths$\;$}
% To aid the discovery process, 
We first establish the connection between ancestral relationships and active causal paths. 
\definecolor{babyblueeyes}{rgb}{0.63, 0.79, 0.95}
\begin{figure}[t]
\begin{tikzpicture}[every node/.style={circle, draw, minimum size=0.3cm, fill=babyblueeyes, fill opacity=1}]
\useasboundingbox (-1, -1.5) rectangle (12, 1);
% Define spacing between diagrams
\def\columnSpacing{3.5} % horizontal spacing between centers of diagrams
% Diagram 1: Two nodes with no edges
\node (A1) at (0,0) {$x_i$};
\node (B1) at (1.5,0) {$x_j$};
\node[draw=none, fill=none] at (0.75, +0.6) {AP1};

% Diagram 2: One node causing two others with dashed arrows
\node (A2) at (\columnSpacing,0) {$x_k$};
\node (B2) at (\columnSpacing-1,-1) {$x_i$};
\node (C2) at (\columnSpacing+1,-1) {$x_j$};
\draw [dashed, -{Latex[length=2mm]}] (A2) -- (B2);
\draw [dashed, -{Latex[length=2mm]}] (A2) -- (C2);
\node[draw=none, fill=none] at (\columnSpacing, +0.6) {AP2};

% Diagram 3: Two nodes with one dashed arrow
\node (A3) at (2*\columnSpacing-1,0) {$x_i$};
\node (B3) at (2*\columnSpacing+1,0) {$x_j$};
\draw [dashed, -{Latex[length=2mm]}] (A3) -- (B3);
\node[draw=none, fill=none] at (2*\columnSpacing, +0.6) {AP3};

% Diagram 4: Combination of Diagram 2 and 3
\node (A4) at (3*\columnSpacing,0) {$x_k$};
\node (B4) at (3*\columnSpacing-1,-1) {$x_i$};
\node (C4) at (3*\columnSpacing+1,-1) {$x_j$};
\draw [dashed, -{Latex[length=2mm]}] (A4) -- (B4);
\draw [dashed, -{Latex[length=2mm]}] (A4) -- (C4);
\draw [dashed, -{Latex[length=2mm]}] (B4) -- (C4);
\node[draw=none, fill=none] at (3*\columnSpacing, +0.6) {AP4};
% \node[draw=none, fill=none] at (1.5*\columnSpacing+0.5, -2.5); 
\end{tikzpicture}
\caption{Enumeration of active causal path relation types between a pair of nodes $x_i$ and $x_j$. Dashed arrows indicate ancestorship.}
\label{fig:active_causal_path_ancestors}
%\vspace{-10pt}
\end{figure}
We exhaustively enumerate and define the potential pairwise causal ancestral path relations in Figure~\ref{fig:active_causal_path_ancestors} and Lemma~\ref{lemma:causal_path_partition_ancestors} (proof in Appendix~\ref{appendix:causal_path_partition_ancestors}):

\begin{restatable}[Active Causal Ancestral Path Relation Enumeration]{lemma}{causalPathPartition}\label{lemma:causal_path_partition_ancestors}
    Each pair of distinct nodes $x_i,x_j \in V$ can be in one of four possible active causal ancestral path relations: AP1) no active path exists between $x_i,x_j$;
     AP2) there exists an active backdoor path between $x_i,x_j$, but there is \emph{no} active frontdoor path between them;
     AP3) there exists an active frontdoor path between $x_i,x_j$, but there is \emph{no} active backdoor path between them;
      AP4) there exists an active backdoor path between $x_i,x_j$, \emph{and} there exists an active frontdoor path between them.
\end{restatable}

Next, in Lemma~\ref{lemma:path_ances_relation}, we summarize the connection between causal paths and ancestral relationships (proof in Appendix \ref{appendix:path_ances_relation}):
\begin{restatable}{lemma}{LemmaPath}
% \begin{lemma}
\label{lemma:path_ances_relation} 
The ancestral relationship between
a pair of distinct nodes $x_i,x_j \in V$
can be expressed using
% in terms of
active causal path
relations: 
$x_i,x_j$ are not ancestrally related if and only if they are in AP1 or AP2 relation; and $x_i,x_j$ are ancestrally related if and only if they are in AP3 or in AP4 relation.
% \end{lemma}
\end{restatable}
%put something here to signal to the reader the following lemmas will definition the necessary and sufficient conditions (identifiability conditions) overview
The active causal ancestral path relation of a pair of nodes
$x_i, x_j$ 
that are not ancestrally related
can be determined through marginal independence testing and sequential univariate regressions as illustrated in Lemmas~\ref{lemma:AP1} and~\ref{lemma:AP2} (proofs in Appendices~\ref{appendix:AP1},~\ref{appendix:AP2}):
\begin{restatable}[AP1]{lemma}{APOne}
\label{lemma:AP1}
    Vertices $x_i,x_j$ are in AP1 relation if and only if $x_i \ind x_j$. 
\end{restatable}
\begin{restatable}[AP2]{lemma}{APTwo}
\label{lemma:AP2}Let $M$ be the set of mutual ancestors between a pair of vertices $x_i$ and $x_j$, i.e., $M = \textit{An}(x_i)\cap \textit{An}(x_j)$. Let $x^M_i,x^M_j$ be the result of sequentially regressing all mutual ancestors in $M$ out of $x_i,x_j$ with univariate regressions, in any order. Then, let 
$r_i^j$ be the residual of $x^M_j$ regressed on $x^M_i$, and $r_j^i$ be the residual of $x^M_i$ regressed on $x^M_j$. Suppose $x_i \nind x_j$. Then, $x_i,x_j$ are in AP2 relation if and only if ${r_i^j} \ind x^M_i$ and $r_j^i \ind x^M_j$. 
\end{restatable}
If a pair of nodes $x_i,x_j$ is ancestrally related, fully ascertaining their ancestral relation involves discerning between the ancestor and descendent. 
As illustrated in Lemmas \ref{lemma:AP3} and \ref{lemma:AP4} (proofs in Appendices~\ref{appendix:AP3},~\ref{appendix:AP4}), this can be determined through marginal independence testing after sequential univariate regressions with respect to the mutual ancestor set.
\begin{restatable}[AP3]{lemma}{APThree}
\label{lemma:AP3}
Let $r_i^j$ be the residual of the $x_j$ regressed on $x_i$, and $r_j^i$ be the residual of $x_i$ regressed on $x_j$. Vertices $x_i,x_j$ are in AP3 relation if and only if $x_i \nind x_j$ and one of the following holds: 1) $x_i \ind r_i^j$ and $x_j \nind r_j^i$, corresponding to $x_i \in \textit{An}(x_j)$, or 2) $x_i \nind r_i^j$ and $x_j \ind r_j^i$, 
corresponding
to  $x_j \in \textit{An}(x_i)$.
\end{restatable}
\begin{restatable}[AP4]{lemma}{APFour}
\label{lemma:AP4}
   Let $M$, $r_i^j, r_j^i,x^M_i,x^M_j$ be as defined in Lemma \ref{lemma:AP2}. Suppose $x_i,x_j$ not in AP3 relation.
   Then, $x_i,x_j$ are in AP4 relation if and only if $x_i \nind x_j$ and one of the following holds: $r_i^j \ind x^M_i$ and $r_j^i \nind x^M_j$ corresponding to $x_i\in\textit{An}(x_j)$, or 2) $r_i^j \nind x^M_i$ and $r_j^i \ind x^M_j$ corresponding to $x_j\in\textit{An}(x_i)$.
\end{restatable}

\textbf{Linear Hierarchical Topological Sort$\;$}
We propose 
LHTS in Algorithm~\ref{algo:lin} to leverage the above results to discover a hierarchical topological ordering. Marginal independence tests and pairwise regressions are used to identify active causal ancestral path relations between pairs of vertices.

\begin{algorithm}[t]
    \caption{\textbf{LHTS: Linear Hierarchical Topological Sort }}\label{algo:lin}
      \vspace{-5mm}
    \begin{multicols}{2}
    \begin{algorithmic}[1]
        \State \textbf{input:} features $x_1,\ldots,x_{d} \in V$.
        \State \textbf{output:} hierarchical topological sort $\pi_L(V)$.
        \State \textbf{initialize:} ancestral relations set $\mathrm{ARS}$.
            \State \textbf{Stage 1: AP1 Relations}
            \For{all pairs $x_i,x_j$} 
            \If{$x_i \ind x_j$} 
            \State Store $ x_i,x_j$ are not related in $\mathrm{ARS}$.
            \EndIf
        \EndFor
        \State \textbf{Stage 2: AP3 Relations}
        \For{all $x_i,x_j$ with unknown relations}
                \State Set $r_i^j$ as residual of $x_j$ regressed on $x_i$
                \State Set $r_j^i$ as residual of $x_i$ regressed on $x_j$.
    
                \If{$ r_i^j \ind x_i \textbf{~and~} r_j^i \not\ind x_j $} 
                \State Store $ x_i\in\textit{An}(x_j)$  in $\mathrm{ARS}$.\EndIf
                \If{$ r_i^j \not\ind x_i \textbf{~and~} r_j^i \ind x_j $} 
                \State  Store $x_j \in \textit{An}(x_i)$ in $\mathrm{ARS}$.\EndIf
            \EndFor
        \State \textbf{Stage 3: AP2 and AP4 Relations}
        \While{$\exists$ $x_i,x_j$ with unknown relations}
            \For{$x_i, x_j$ with unknown relations}
                \State Sequentially regress $x_i,x_j$ on mutual \hspace*{2.7em} ancestors, store final residuals $r_i^j, r_j^i$.
                
                \If{$ r_{i}^{j} \ind x_{i} \textbf{~and~} \prescript{}{I}{r_j^i}  \ind x_{j} $} 
                \State Store  $x_i,x_j$ are not related in \hspace*{4.5em}$\mathrm{ARS}$.
                \ElsIf{$ r_i^j \ind x_{i} \textbf{~and~} r_j^i \not\ind x_{j} $} 
                \State Store $x_i \in \textit{An}(x_j)$ in $\mathrm{ARS}$.
                \ElsIf{$ r_i^j \nind x_{i} \textbf{~and~} r_j^i  \ind x_{j} $} 
                \State   Store $x_j \in \textit{An}(x_i)$ in $\mathrm{ARS}$.
                \EndIf
            \EndFor
        \EndWhile
\State \textbf{Stage 4: Obtain sort by subroutine \textbf{AS}}
\State \textbf{return:} $\pi_L(V)\gets {\textbf{AS}}(\text{ARS})$ 
    \end{algorithmic}
    \end{multicols}
      \vspace{-10pt}
\end{algorithm}

LHTS discovers all pairwise active causal ancestral path relations: Stage 1 discovers all AP1 relations using marginal independence tests; Stage 2 discovers all AP3 relations by detecting when pairwise regressions yield an independent residual in only one direction; Stage 3 iteratively discovers all AP2 and AP4 relations through marginal independence tests and pairwise sequential regressions involving mutual ancestors. Note that Stage 2 can be viewed as a special case of Stage 3, where the mutual ancestor set is empty. The process utilizes proofs provided in Appendices \ref{appendix:AP1}, \ref{appendix:AP3}, \ref{appendix:AP2}, and \ref{appendix:AP4}, respectively. Stage 4 uses the complete set of ancestral relations to build the final hierarchical topological sort via the subroutine \textbf{AS} (Algorithm~\ref{alg:as} in Appendix \ref{appendix:ancestor_sort})
by recursively peeling off vertices with no unsorted ancestors. We show the correctness of Algorithm \ref{algo:lin} in Theorem \ref{theorem:lin_alg} (proof in Appendix \ref{appendix:LHTS}) and subroutine $AS$ (proof in Appendix \ref{appendix:ancestor_sort}), as well as the worst case time complexity in Theorem \ref{theorem:lin_alg_time} (proof in Appendix \ref{appendix:lhts_time}). We provide a walk-through of %Algorithm \ref{algo:lin} 
LHTS in Appendix \ref{appendix:lhts_walk}.
\begin{theorem}\label{theorem:lin_alg}
Given a graph $G$, Algorithm \ref{algo:lin} asymptotically finds a correct hierarchical sort of $G$.
\end{theorem}
%The main contribution to complexity in LHTS comes from running marginal independence tests that are nonparametric with respect to the noise distribution.
%The complexity depends on the specific independence test chosen by user: we use a HSIC-based independence test \citep{gretton_kernel_2007}, which runs in $O(n^2)$.
\begin{theorem}\label{theorem:lin_alg_time}
Given $n$ samples of $d$ vertices generated by a LiNGAM, the worst case runtime complexity of Algorithm \ref{algo:lin} is upper bounded by $O(d^3n^2)$.
\end{theorem}

%The worst case complexity occurs when the graph is fully connected, as there are $d$ iterations of Stage 3, each running $O(d^2)$ many marginal independence tests: this leads to a worst case time complexity of $O(d^3n^2)$. 

LHTS can be seen as a generalization of DirectLiNGAM \citep{shimizu2011directlingam}, where they recursively identify root nodes as vertices that have either AP1 or AP3 relations with every remaining vertex at each step.
% in each step of their algorithm. 
The next lemma, 
% we show that
Lemma~\ref{lemma:root_vertex} (proof in Appendix \ref{lemma:roots}) implies that LHTS discovers all
% , so
root vertices in Stage 2, obtaining the first layer in the hierarchical sort. LHTS extends DirectLiNGAM by discovering AP2 and AP4 relations, allowing us to recover a compact hierarchical sort.

\begin{restatable}{lemma}{rootVertex}\label{lemma:root_vertex}A vertex is a root vertex if and only if it has AP1 or AP3 relations with all other vertices.
\end{restatable}
%%%%%%%%%%%%%%%%%%%%%%%%%%
%\vspace{-5pt}
\section{Nonlinear Setting}\label{sec:nonlin_setting}
%\vspace{-3pt}
In this section, we develop a version of Algorithm~\ref{algo:lin} for ANMs that feature only nonlinear causal functions $f$.
%the nonlinear ANM(Eq. \eqref{eq:ANM}) setting.
Rather than detecting ancestor-descendant relationships, we outline the connection between active causal paths and parent-child relationships.
We  assume the unique identifiability of the nonlinear ANM as described by Peters et al. \citep{peters_causal_2014}, and provide the conditions
 in Appendix \ref{appendix:nonlin_identify}. 
 
\definecolor{babyblueeyes}{rgb}{0.63, 0.79, 0.95}
\begin{figure}[t]
\begin{tikzpicture}[every node/.style={circle, draw, minimum size=0.3cm, fill=babyblueeyes, fill opacity=1}]
\useasboundingbox (-1, -1.5) rectangle (12, 1);
% Define spacing between diagrams
\def\columnSpacing{3.5} % horizontal spacing between centers of diagrams
% Diagram 1: Two nodes with no edges
\node (A1) at (0,0) {$x_i$};
\node (B1) at (1.5,0) {$C$};
\node (C1) at (.75,-1) {$x_j$};
\draw [solid, -{Latex[length=2mm]}] (B1) -- (C1);
\node[draw=none, fill=none] at (0.75, +0.6) {PP1};

% Diagram 2: One node causing two others with dashed arrows
\node (A1) at (\columnSpacing,0) {$x_i$};
\node (B1) at (\columnSpacing+1.5,0) {$C$};
\node (C1) at (\columnSpacing+.75,-1) {$x_j$};
\draw [solid, -{Latex[length=2mm]}] (A1) -- (C1);
\draw [solid, -{Latex[length=2mm]}] (B1) -- (C1);
\node[draw=none, fill=none] at (\columnSpacing+0.75, +0.6) {PP2};

% Diagram 3: Two nodes with one dashed arrow
\node (A3) at (2*\columnSpacing,0) {$x_i$};
\node (B3) at (2*\columnSpacing+1.5,0) {$C$};
\node (C3) at (2*\columnSpacing+.75,-1) {$x_j$};
\draw [solid, -{Latex[length=2mm]}] (B3) -- (C3);
\draw [solid, -{Latex[length=2mm]}] (A3) -- (C3);
\draw [dashed] (A3) -- (B3);
\node[draw=none, fill=none] at (2*\columnSpacing+0.75, +0.6) {PP3};

% Diagram 3: Two nodes with one dashed arrow
\node (A3) at (3*\columnSpacing,0) {$x_i$};
\node (B3) at (3*\columnSpacing+1.5,0) {$C$};
\node (C3) at (3*\columnSpacing+.75,-1) {$x_j$};
\draw [solid, -{Latex[length=2mm]}] (B3) -- (C3);
\draw [dashed] (A3) -- (B3);
\node[draw=none, fill=none] at (3*\columnSpacing+0.75, +0.6) {PP4};
% \node[draw=none, fill=none] at (1.5*\columnSpacing+0.5, -2.5); 
\end{tikzpicture}
\caption{Enumeration of the potential active causal paths among a fixed variable $x_j$, one of its potential parents $x_i$, and $C=\text{PA}(x_j)\setminus{x_i}$.
Solid arrows denote
parenthood relations, and
undirected dashed connections indicate the existence of
active paths.
}
\label{fig:active_causal_path_parents}
%\vspace{-10pt}
\end{figure}
\textbf{Nonlinear Topological Sort$\;$}
In the linear setting, we determined ancestral relations through a sequence of pairwise regressions that led to independent residuals. 
However, a naive extension of this method into the nonlinear setting would fail, as regressions yield independent residuals under different conditions in the nonlinear case. For clarity, we demonstrate how LHTS fails to correctly recover causal relationships in an exemplary 3-node DAG with nonlinear causal mechanisms.

Consider a DAG $G$ with three vertices $x_1,x_2,x_3$, where $x_1 \rightarrow x_3, x_2 \rightarrow x_3$. The functional causal relationships are nonlinear, given by $x_1 = \varepsilon_1, x_2 = \varepsilon_2, x_3 = x_1x_2 +\varepsilon_3$, where $\varepsilon_i$ are mutually independent noise variables. We focus on whether LHTS can recover the parent-child relationship between $x_1$ and $x_3$. LHTS finds that the relationship between $x_1,x_3$ is unknown in Stage 1. In Stage 2, LHTS runs pairwise regressions between $x_1,x_3$ but \emph{incorrectly concludes that $x_1,x_3$ are not in AP3 relation} because neither pairwise regression provides an independent residual; both parents of $x_3$ must be included in the covariate set for an independent residual to be recovered.

%because the presence of any indirect active causal paths induces dependent residuals from pairwise regression due to omitted variable bias \citep{pearl_causal_2009}, even when regressing descendants on ancestors.

To handle nonlinear causal relationships, we shift our focus to searching for a different set of local substructures: the connection between active causal parental paths and the existence of parent-child relationships. We will use the existence of specific parent-child relationships to first obtain a superset of root vertices, then prune away non-roots. Once all root vertices are identified, we build the hierarchical topological sort through nonparametric regression, layer by layer.

Given a
vertex $x_j$ and one of its potential parents $x_i$, 
we first provide an enumeration of all potential active casual parental path types between them with respect to $C=\text{PA}(x_j)\setminus{x_i}$  (which could potentially be the empty set) in Figure~\ref{fig:active_causal_path_parents} and Lemma \ref{lemma:causal_parent_path_partition} (proof in Appendix~\ref{appendix:causal_parent_path_partition}).

\begin{restatable}[Active Causal Parental Path Relation Enumeration]{lemma}{parentenum}\label{lemma:causal_parent_path_partition}
Let $x_i,x_j \in V$ be a pair of distinct nodes, where $x_i$ is one of the potential parents of $x_j$.  
    Let $C = \text{PA}(x_j)\setminus{x_i}$. 
    There are in total four possible types of active causal parental path relations between $x_i$ and $x_j$ with respect to $C$: 
    PP1) $x_i$ and $x_j$ are \emph{not} directly causally related, and \emph{no} active path exists between $x_i$ and $C$;
    PP2) $x_i$ directly causes $x_j$ ($x_i\to x_j$), and \emph{no} active path exists between $x_i$ and $C$;
     PP3) $x_i$ directly causes $x_j$ ($x_i\to x_j$), and there exists an active path between $x_i$ and $C$;
      PP4) $x_i$ and $x_j$ are \emph{not} directly causally related, and there exists an active path between $x_i$ and $C$.
\end{restatable}
Next, for a pair of distinct vertices $x_i, x_j\in V$, we 
establish the connection between pairwise independence properties and active causal parental path relations in Lemma~\ref{lemma:PP1} (proof in Appendix~\ref{appendix:PP1}). This 
allows us to
reduce the 
cardinality of the potential pairs of vertices
under consideration
in the later stages of the algorithm.

\begin{restatable}[Non-PP1]{lemma}{nonppone}\label{lemma:PP1}
Vertices $x_i,x_j\in V$ are not in PP1 relation if and only if $x_i \nind x_j$.
\end{restatable}

In Lemma~\ref{lemma:PP2}, we show that 
all pairs of vertices that are in PP2 relation can be identified through local nonparametric regressions (proof in \ref{appendix:PP2}).

\begin{restatable}[PP2]{lemma}{pptwo}\label{lemma:PP2}
    Let $x_i,x_j\in V$, $P_{ij} = \{x_k\in V: x_k\ind x_i, x_k\not\ind x_j\}$, $r_i^j$ be the residual of $x_j$ nonparametrically regressed on $x_i$, and $r^j_{i,P}$ be the residual of $x_j$ nonparametrically regressed on $x_i$ and all $x_k \in P_{ij}$.
    Suppose $x_i$ and $x_j$ are not in PP1 relation. Then, $x_i$ and  $x_j$ are in PP2 relation if and only if one of the following holds: 1) $x_i \ind r_i^j$ or 2) $x_i \ind r^j_{i,P}$.
\end{restatable}
Condition 1) of Lemma \ref{lemma:PP2} is relevant when $C=\emptyset$: in this case, pairwise regression identifies $x_i$ as a parent of $x_j$. Condition 2) is relevant when $C\neq \emptyset$: we leverage the independence of $x_i$ from the rest of $x_j$'s parents 
to generate $P_{ij}$, a set that contains $C$, but does not contain any of $x_j$'s descendants. 
If an independent residual were to be recovered by
nonparametrically regressing $x_j$ onto $x_i$ and $P_{ij}$, we identify $x_i$ as a parent of $x_j$. 

Let $W$ be the set of all parent vertices that are in PP2 relation with at least one vertex, i.e., the union of $x_i$ satisfying either condition of Lemma \ref{lemma:PP2}. We now show that all non-isolated root vertices are contained in $W$, and they can be differentiated from non-roots in $W$ that are also in PP2 relations.
\begin{restatable}[Roots]{lemma}{rootlemma}\label{lemma:root_identification_nonlinear}
   All non-isolated root vertices are contained in $W$. In addition, $x_i \in W$ is a root vertex if and only if 
   1) $x_i$ is 
   not a known descendant of 
   any $x_j \in W$, 
   % \kyra{should W include both pairs or just the parent of the pair?} okay
   % as a child \kyra{condition 2) is a bit unclear, what do we mean by not in PP2 as a child?},
   and 2)  
   for each $x_j \in W$, 
   either a) $x_i \ind x_j$, b) $x_j$ is in PP2 relation to $x_i$, i.e., $x_j \in \text{Ch}(x_i)$, or c)
   there exists a child of $x_i$, denoted by $x_k$, that cannot be d-separated from $x_j$ given $x_i$,
i.e., $x_j \nind x_k |x_i$.
\end{restatable}

Lemma \ref{lemma:root_identification_nonlinear}
relies on the following intuition: for any non-isolated root vertex $x_i$ and descendant $x_k \in \text{De}(x_i)$, there exists a child of $x_i$ that 1) lies on a directed path between $x_i$ and $x_k$, and 2) is in PP2 relation with $x_i$. Vertex $x_i$ will fail to d-separate this specific child from the marginally dependent non-root descendant $x_k$.\footnote{Vertices $x_i, x_j$ are said to be d-separated by a set $\textbf{Z}$ iff there is no active path between $x_i,x_j$ relative to $\textbf{Z}$.} On the other hand, non-roots in $W$ 
must d-separate the children they are in PP2 relation to from all marginally dependent roots: this asymmetry allows non-roots to be pruned. See Appendix \ref{appendix:root_id_nonlinear} for a detailed proof.

We propose a method, Algorithm~\ref{algo:nhts}, that leverages the above results to discover a hierarchical topological ordering: we first
use the above lemmas to identify the roots, then use nonparametric regression to discover the rest of the hierarchical sort. 
\begin{algorithm}[t]
    \caption{\textbf{NHTS}: Nonlinear Hierarchical Topological Sort}
    \label{algo:nhts}
      \vspace{-5mm}
    \begin{multicols}{2}
    \begin{algorithmic}[1]
        \State \textbf{input:} vertices $x_1,\ldots,x_{d} \in V$.
        \State \textbf{output:} hierarchical sort $\pi_L(V)$.
        \State \textbf{initialize:} parent relations set $\mathrm{PRS}$.
            \State \textbf{Stage 1: Not-PP1 Relations}
            \For{all pairs $x_i,x_j$} 
            \If{$x_i \nind x_j$} 
            \State $\mathrm{PRS}: x_i,x_j$ not in PP1 relations.
            \EndIf
        \EndFor
        \If{$x_i$ is in PP1 relation with all vertices} 
        \State PRS: $x_i$ is isolated, sort $x_i$: $\pi_L(x_i)=0$.
        \EndIf
        \State \textbf{Stage 2: PP2 Relations}
        \For{all $x_i,x_j$ not in PP1 relations}
                \State Set $r_i^j$ as residual of $x_j$ regressed on $x_i$.
                \State Set $r^j_{i,P}$ as residual of $x_j$ regressed on \hspace*{1.7em}$x_i, P_{ij}$.
                \If{$ x_i \ind r_i^j  \textbf{~or~} x_i \not\ind r^j_{i,P} $} 
                \State $\mathrm{PRS}:x_i,x_j \text{ are in PP2 relations, }x_i$
                \hspace*{3em}$\in\textit{Pa}(x_j)$.
                \EndIf
            \EndFor
        \State \textbf{Stage 3: Root Identification}
        \For{$x_i \in$ PP2 relation}
        \State Check if $x_i$ does not d-separate any child \hspace*{1.4em}from all marginally dependent vertices. If \hspace*{1.3em} so, then update $\mathrm{PRS}: x_i$ is a root, sort
         \hspace*{1.3em} $x_i$ into layer 0, $\pi_L(x_i)=0$.
        \EndFor
        \State \textbf{Stage 4: Obtain Sort}
        \For{$k \in \{1,\ldots,d-1\}$}
        \For{all unsorted $x_i$}
        \State Set $r_i$ as the residual of $x_i$ regressed \hspace*{3em}on sorted features in $\pi_H$.
         \State If $r_i \ind x_j \forall x_j \in\pi_H$, add $x_i$ into \hspace*{3em}the current layer $\pi_L(x_i)=k$.
        \EndFor
        \EndFor
\State \Return $\pi_L(V)$ 
    \end{algorithmic}
    \end{multicols}
      \vspace{-3mm}
\end{algorithm}

In Stage 1, we discover all pairs $x_i,x_j$ not in PP1 relation by testing for marginal dependence; in Stage 2, we leverage Lemma \ref{lemma:PP2} to find the vertex pairs that are in PP2 relations, a superset of the root vertices; in Stage 3 we prune non-roots by finding they d-separate their children from at least one dependent vertex in $W$ (Lemma \ref{lemma:root_identification_nonlinear}); in Stage 4 we identify vertices in the closest unknown layer by regressing them on sorted nodes and finding independent residuals. We show the correctness of Algorithm \ref{algo:nhts} in Theorem \ref{theorem:nonlin_alg} (proof in Appendix \ref{appendix:NLHTS_proof}), and the worst case time complexity in Theorem \ref{theorem:nonlin_alg_time} (proof in Appendix \ref{appendix:nhts_time}). We provide a walk-through of Algorithm \ref{algo:nhts} in Appendix \ref{appendix:nhts_walk}.

\begin{restatable}{theorem}{algotwocorrect}\label{theorem:nonlin_alg}
Given a graph $G$, Algorithm \ref{algo:nhts} asymptotically finds a correct hierarchical sort of $G$.
\end{restatable}
\begin{restatable}{theorem}{algotworun}\label{theorem:nonlin_alg_time}
Given $n$ samples of $d$ vertices generated by a identifiable nonlinear ANM, the worst case runtime complexity of Algorithm \ref{algo:nhts} is upper bounded by $O(d^3n^3)$.
\end{restatable}
 
 %NHTS yields a correct hierarchical sort with far fewer high-dimensional regressions than RESIT or NoGAM: for example, to find the last vertex in the topological sort, these methods run $d$ number of $(d-1)$-dimensional regressions, while NHTS runs only two $(d-2)$-dimensional regressions. 
 The number of nonparametric regressions run by NHTS in each step is actually inversely related to the size of the covariate sets, while the number of regressions in each step of RESIT and NoGAM are directly proportionate to the covariate set size. We provide a formal analysis of the reduction in complexity for the worst case (fully connected DAG) in Theorem \ref{nhts_time_red} (proof in Appendix \ref{appendix:nhts_time_red}):

 \begin{restatable}{theorem}{nhtsrunred}\label{nhts_time_red}
     Consider a fully connected DAG $G=(V,E)$
with nonlinear ANM. Let $d:= |V|$. 
Let $n^\mathrm{NHTS}_k$ be the number of nonparametric regressions with covariate set size $k\in[d-2]$ run by NHTS when sorting $V$; we similarly define $n^\mathrm{RESIT}_k$ and $n^\mathrm{NoGAM}_k$ respectively.  Then, $n^\mathrm{NHTS}_k = d - k$, and $ n^\mathrm{RESIT}_k = n^\mathrm{NoGAM}_k = k+1$. This implies that for all  $k > \frac{d}{2}$, $n^\mathrm{NHTS}_k<n^\mathrm{RESIT}_k = n^\mathrm{NoGAM}_k$.
 \end{restatable}

%\vspace{-5pt}
\section{Edge Discovery}\label{sec:edge_discovery}
%\vspace{-3pt}
Parent selection from a topological ordering $\pi$ via regression is 
traditionally
a straightforward task in the infinite sample setting: for each vertex $x_i$, $\pi$ establishes $J_i=\{x_k: \pi(x_k)<\pi(x_i)\}$, a superset of $x_i$'s parents that contains none of $x_i$'s descendants. A general strategy for pruning $\text{Pa}(x_i)$ from $J_i$ is to regress $x_i$ on $J_i$ and check which $x_k \in J_i$ are relevant predictors. The key issue is that $J_i$ grows large in high-dimensional graphs: current edge pruning methods either make strong parametric assumptions or suffer in sample complexity. 
Lasso and GAM methods impose linear and additive models, failing to correctly identify parents in highly nonlinear settings. RESIT assumes a more general nonlinear ANM, but requires huge sample sizes and oracle independence tests for accurate parent set identification: authors in \citep{peters_causal_2014} confirm this, saying "despite [our] theoretical guarantee[s], RESIT does not scale well to a high number of nodes."

%The specifics of how the relevance of predictors is determined depends on the regression method used: Lasso regression identifies the set of covariates with nonzero coefficients, Generalized Additive Models identify the set of covariates with smooth functions significantly different from 0, and RESIT identifies coviarates necessary for residual independence.

%Authors of RESIT propose a nonparametric approach to checking if $x_j\in \text{Pa}(x_i)$ by regressing $x_i$ on $J_i\setminus{x_j}$; the residual is dependent on $x_j$ if and only if $x_j \in \text{Pa}(x_i)$
%\citep{peters_causal_2014}. However, this approach relies on high-dimensional-non parametric regressions, running $d$ $d-2$-dimensional regressions in the first step. Additionally, a regression based approach suffers from model-fitting issues. The authors \citep{peters_causal_2014} confirm this, saying "despite [our] theoretical gaurantee[s], RESIT does not scale well to a high number of nodes", due to finite sample issues.

%In finite sample settings, parametric methods such as Lasso regression or covariate hypothesis testing with generalized additive models are used: $x_i$ is regressed on all of $J_i$, and $x_j\in J_i$ that have nonzero coefficients are determined as parents of $x_i$. 

We propose an entirely nonparametric constraint-based method that uses a local conditioning set $Z_{ij}$ to discover whether $x_i\in \text{Pa}(x_j)$, rather than $J_i$, outperforming previous methods by relaxing parametric assumptions and conditioning on fewer variables. The following lemma outlines a sufficient condition for determining whether an edge exists between two vertices (proof in Appendix \ref{appendix:edge_lemma}), visualized in Figure \ref{fig:edge_DAG}.
%\vspace{-10mm}
\begin{figure}[t!]
\centering
\begin{tikzpicture}[scale=0.1, every node/.style={inner sep=0pt}]
\definecolor{mycolor1}{HTML}{9e194d}
\definecolor{mycolor2}{HTML}{baa600}
\definecolor{mycolor3}{HTML}{96bfe6}

% Nodes
\draw [fill=mycolor1, fill opacity=0.3] (0,0) circle (3); % Z_1
\draw (0,0) node {$C$};

\draw [fill=mycolor2, fill opacity=0.2] (15,0) circle (3); % X
\draw (15,0) node {$x_i$};

\draw [fill=mycolor1, fill opacity=0.3] (30,0) circle (3); % X
\draw (30,0) node {$M$};

\draw [fill=mycolor3, fill opacity=0.2] (45,0) circle (3); % Y
\draw (45,0) node {$x_j$};

% Arrows
% Z_1 to X
\draw [black, -{Latex[length=2mm,width=4pt]}] (3,0) -- (12,0);

% Z_1 to Y (curving around X)
\draw [red, -{Latex[length=2mm,width=4pt]}] (15,3) to[bend left=45] (42,0);

% X to Y
\draw [dashed, -{Latex[length=2mm,width=4pt]}] (18,0) -- (27,0);

% newwwww
\draw [black, -{Latex[length=2mm,width=4pt]}] (3,0) to[bend left=25] (31,7.3);

\draw [black, -{Latex[length=2mm,width=4pt]}] (33,0) -- (42,0);
\end{tikzpicture}
\caption{DAG corresponding to Lemma \ref{lemma:edge}, which tests whether $x_i \in \text{Pa}(x_j)$ (i.e., whether the red arrow exists).}
\label{fig:edge_DAG}
\vspace{-5pt}
\end{figure}

\begin{restatable}[Parent Discovery]{lemma}{parentedge}\label{lemma:edge}
Let $\pi$ be a topological ordering, $x_i,x_j$ such that $\pi(x_i)<\pi(x_j)$. Let $Z_{ij} = C_{ij} \cup M_{ij} $, where
  $ C_{ij} = \{x_k: x_k \in \text{Pa}(x_i), x_k \nind x_j\}, M_{ij} = \{x_k: x_k \in \text{Pa}(x_j), \pi(x_i) < \pi(x_k) < \pi(x_j)\}.$
Then, $x_i \rightarrow x_j \iff x_i \not \ind x_j |Z_{ij}$.
\end{restatable}
%\vspace{-2pt}
The intuition is that to determine whether $x_i\rightarrow x_j$, instead of conditioning on all potential ancestors of $x_j$, it suffices to condition on potential confounders of $x_i,x_j$ ($C_{ij}$) and potential mediators between $x_i$ and $x_j$ ($M_{ij}$). This renders all backdoor and frontdoor paths inactive, except the frontdoor path corresponding to a potential direct edge from $x_i$ to $x_j$.

\textbf{Edge Discovery$\;$} We propose an algorithm that leverages the above results to discover the true edges admitted by any topological ordering by running the described conditional independence test in a specific ordering. We give the implementation for pruning a topological sort here, but our approach can be generalized to a hierarchical version (see Appendix \ref{appendix:hierarchical_edge}). We show the correctness of Algorithm \ref{algo:edge} in Theorem \ref{theorem:edge_discovery} (proof in Appendix \ref{appendix:edge_discovery}).
\begin{restatable}{theorem}{edcorrect}\label{theorem:edge_discovery}
Given a correct topological ordering of $G$, Algorithm \ref{algo:edge} asymptotically finds correct parent sets $\text{PA}(x_i), \forall x_i \in G$.
\end{restatable}
%\vspace{-2pt}
        \begin{algorithm}[t]
        \caption{\textbf{ED}: Edge Discovery}\label{algo:edge}
            \vspace{-5mm}
             \begin{multicols}{2}
            \begin{algorithmic}[1]
                \State \textbf{input:} features $x_1,\ldots,x_{d} \in V$, topological order $\pi$.
                \State \textbf{initialize:} empty parent sets $\{\text{Pa}(x_i)\}^d_{i=1}$.
                \State \textbf{Step 1: Loop over $\pi$}
                \For{$j \in \{2,\ldots,d\}$}
                \State Set $x_j$ equal to $j^{th}$ index of  $\pi$.
                    \For{$i \in \{j-1,\ldots,1\}$}
                    \State Set $x_i$ equal to $i^{th}$ index of  $\pi$.
                    \State Set $Z_{ij} = \text{Pa}(x_i)\cup \text{Pa}(x_j)$.
                      \State \textbf{Step 2: Test for Parenthood}
                        \If{$x_i \nind x_j | Z_{ij}$}
                            \State Update $x_i \in \text{Pa}(x_j)$.
                        \EndIf
                    \EndFor
                \EndFor
                \State \Return $\{\text{Pa}(x_i)\}^d_{i=1}$.
            \end{algorithmic}
            \end{multicols}
            \vspace{-8pt}
        \end{algorithm}

The key insight is to check each potential parent-offspring relation using the conditional independence test $x_i \ind x_j|Z_{ij}$ such that previous steps in the algorithm obtain all vertices in both $C_{ij}$ and $M_{ij}$. We first fix a vertex $x_j$ whose parent set we want to discover. We then check if vertices ordered before $x_j$ are parents of $x_j$ in reverse order,
starting with the vertex immediately previous to $x_j$ in $\pi$. This process starts at the beginning of $\pi$, meaning we discover parent-offspring relations from root to leaf (see Appendix \ref{appendix:edge_discovery} for a detailed walk-through and proof).
We show the worst case time complexity of Algorithm \ref{algo:edge} in Theorem \ref{theorem:edge_discovery_time} (proof in Appendix \ref{appendix:edge_discovery_time}).
\begin{restatable}{theorem}{edruntime}\label{theorem:edge_discovery_time}
Given $n$ samples of $d$ vertices generated by a model corresponding to a DAG $G$, the runtime complexity of ED is upper bounded by $O(d^2n^3)$.
\end{restatable}

%\emph{Edge discovery} checks $d^2$ possible edges allowed by $\pi_L$ and uses one conditional independence test per check, resulting in a complexity of $O(d^2)$ in the number of conditional independence tests run. The complexity in terms of $n$ will vary depending on the conditional independence test chosen by the user. We use a nonparametric kernel-based conditional independence test \citep{zhang_kernel-based_2011} with $O(n^3)$ complexity in our experiments, leading to an overall $O(d^2n^3)$ time complexity.
%Note that ED improves heavily upon the exponential worst-case complexity of %traditional constraint-based methods like
%PC \citep{glymour_review_2019}.

%\vspace{-5pt}
\section{Experiments}\label{sec:empirical}
\begin{figure}[h!]
%35
    \centering
    \begin{subfigure}[h!]{0.35\textwidth}
        \includegraphics[width=\textwidth]{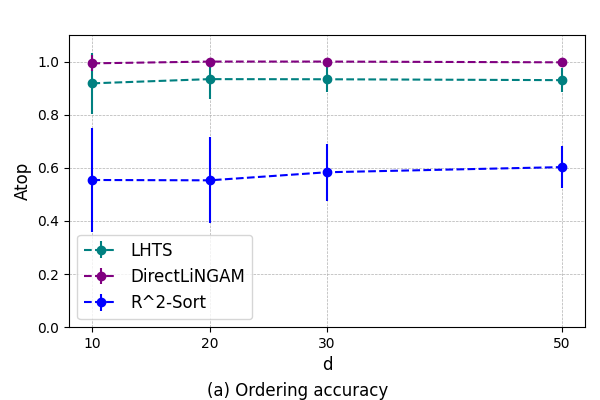}
    \end{subfigure}
    \begin{subfigure}[h!]{0.35\textwidth}
        \includegraphics[width=\textwidth]{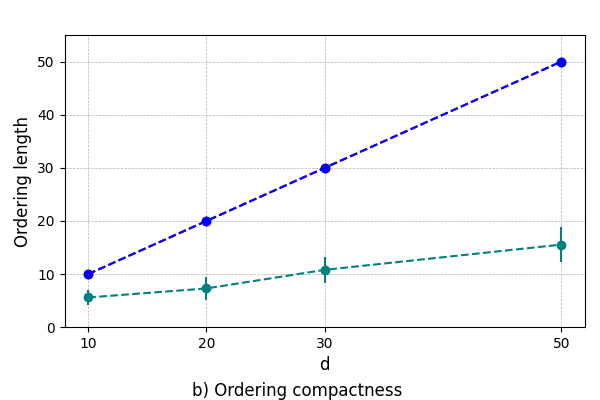}
    \end{subfigure}\\
    \begin{subfigure}[h!]{0.35\textwidth}
        \includegraphics[width=\textwidth]{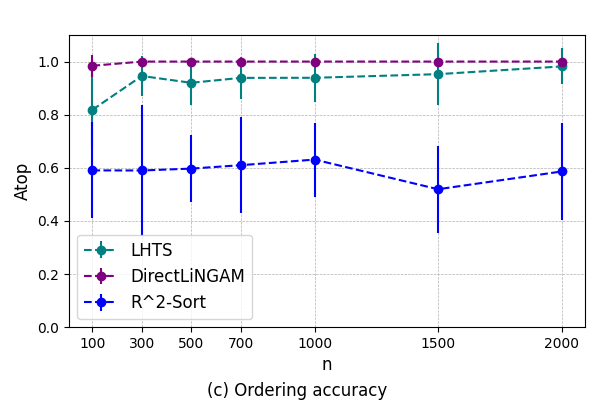}
    \end{subfigure}
    \begin{subfigure}[h!]{0.35\textwidth}
        \includegraphics[width=\textwidth]{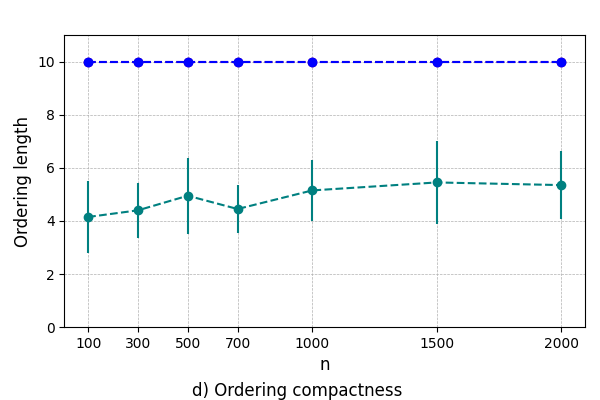}
    \end{subfigure}
    % \vspace{-5pt}
    \caption{Performance of LHTS on synthetic data. Top row:
    % The two graphs (a,b) on the left correspond to
    % trials with 
    $n=500$ with varying dimension $d$. 
    Bottom row: 
    % The two graphs on the right correspond to trials with 
    $d=10$ with varying sample size $n$. See Appendix \ref{appendix: lhtsruntime} for runtime results.}
    \label{fig: lin_sort}
    %\vspace{-15pt}
\end{figure}

\begin{figure}[h]
    \centering
    \begin{subfigure}[b]{0.32\textwidth}
        \includegraphics[width=\textwidth]{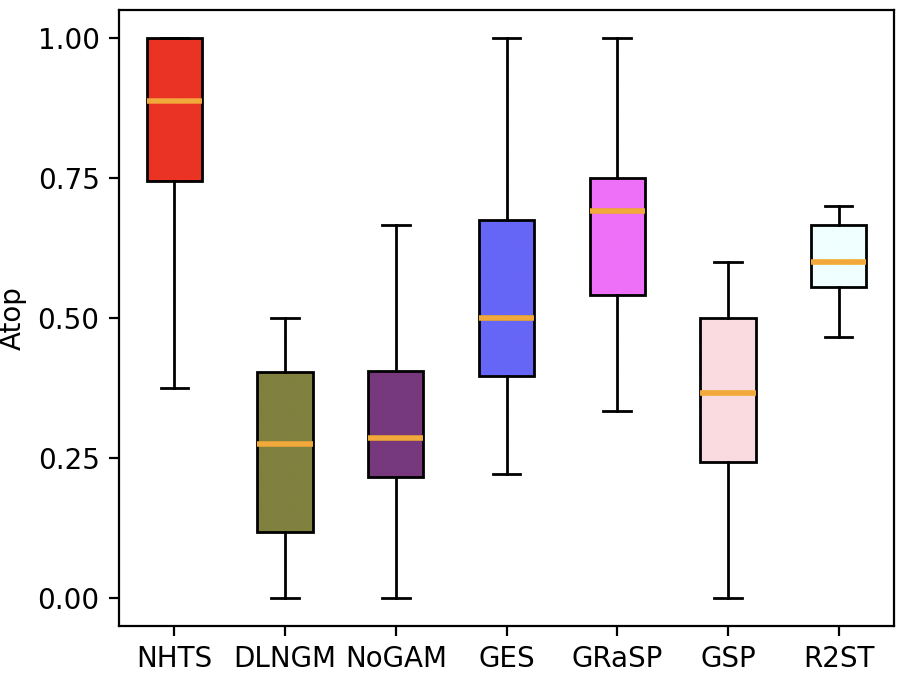}%{figures/GER11.png}
        %{figures/nhts2.png}
    \end{subfigure}
    \begin{subfigure}[b]{0.32\textwidth}
        \includegraphics[width=\textwidth]{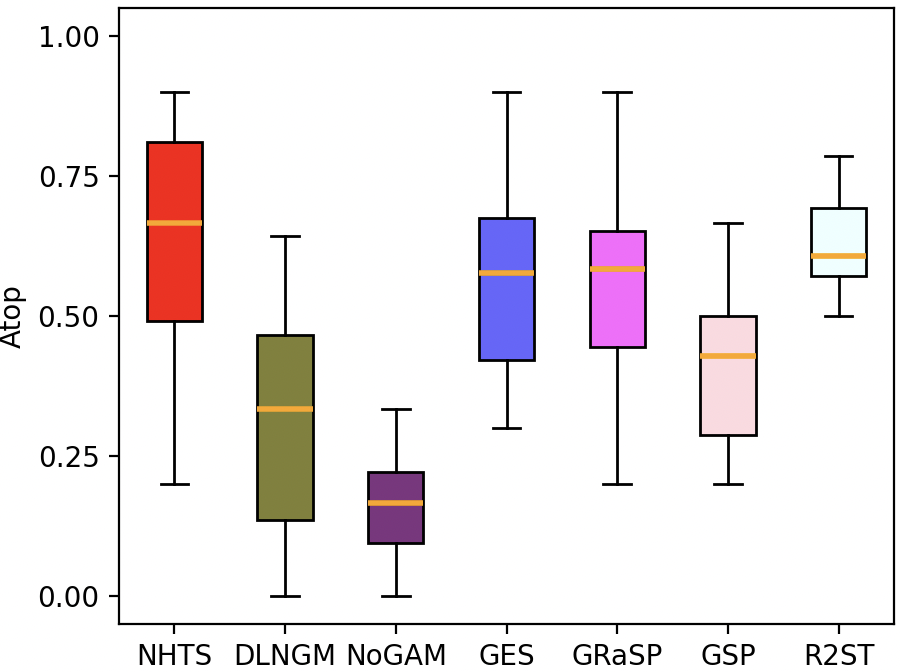}%{figures/LER11.png}
        %{figures/nhts3.png}
    \end{subfigure}
    \begin{subfigure}[b]{0.32\textwidth}
        \includegraphics[width=\textwidth]{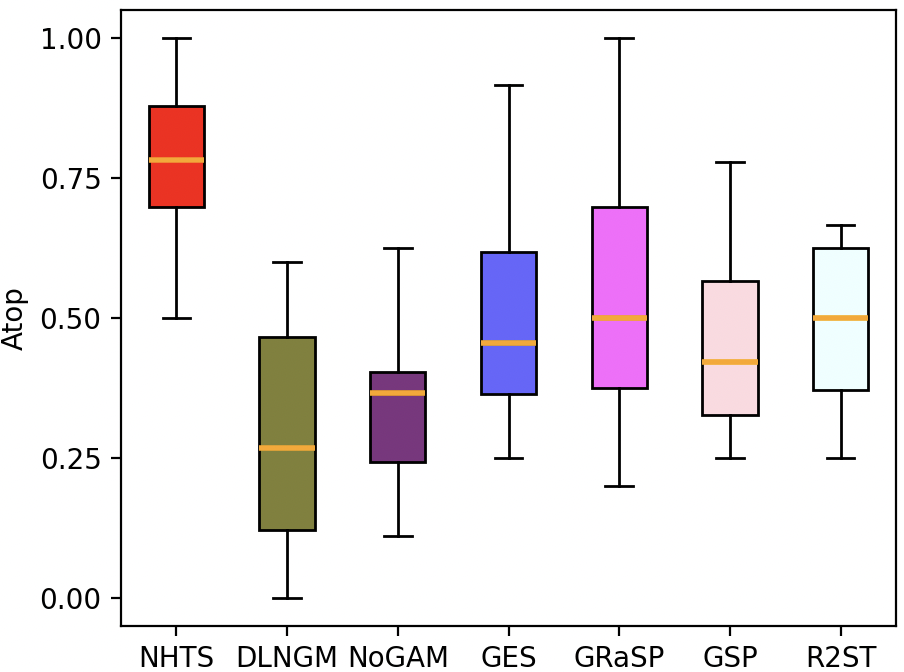}%{figures/UER11.png}
        %{figures/nhts1.png}
    \end{subfigure}
    %\vspace{-5pt}
        \caption{Performance of NHTS on synthetic data, $n=300$, dimension $d=10$, with varying error distributions: Gaussian, Laplace, Uniform (left, middle, right). %Data generated with (left) Gaussian, (middle) Uniform, and Laplacian (right) noise. 
    See Appendix \ref{appendix: nonlin runtimes} for runtime results.} 
    \label{fig: nonlin_sort}
    %\vspace{-5pt}
\end{figure}

\begin{figure}[h!]
    \centering
    \begin{subfigure}[b]{0.32\textwidth}
        \includegraphics[width=\textwidth]%{figures/resitedgedis.png}
        {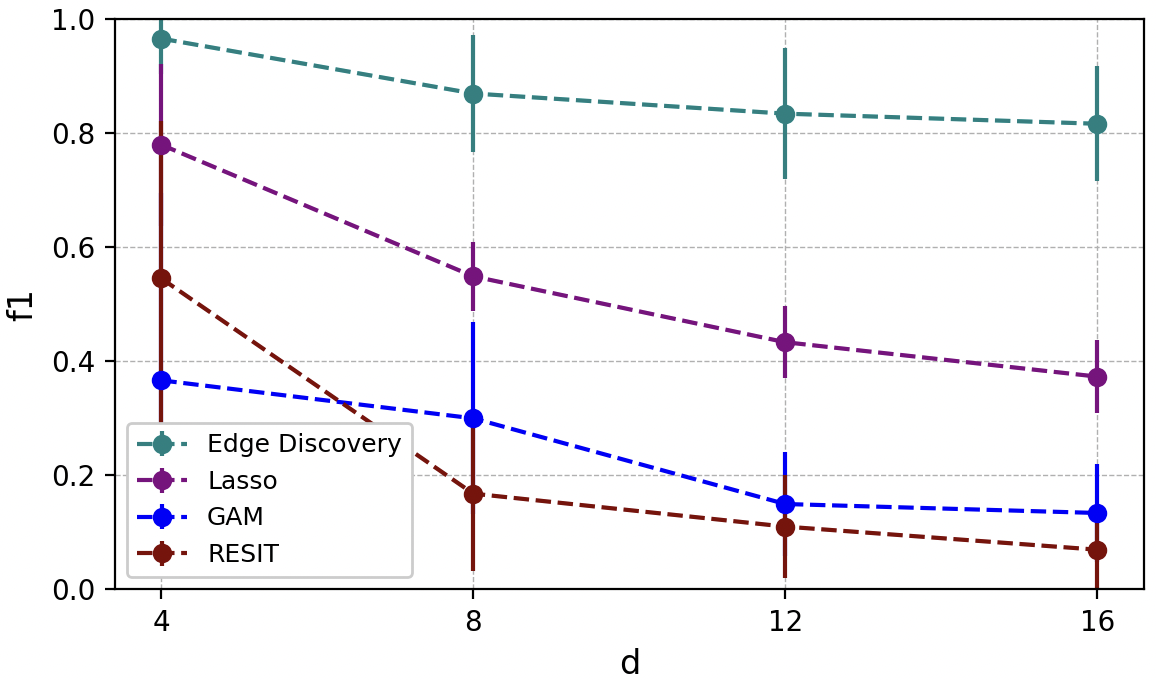}
    \end{subfigure}
    \begin{subfigure}[b]{0.32\textwidth}
        \includegraphics[width=\textwidth]%{figures/holyed1.png}
        {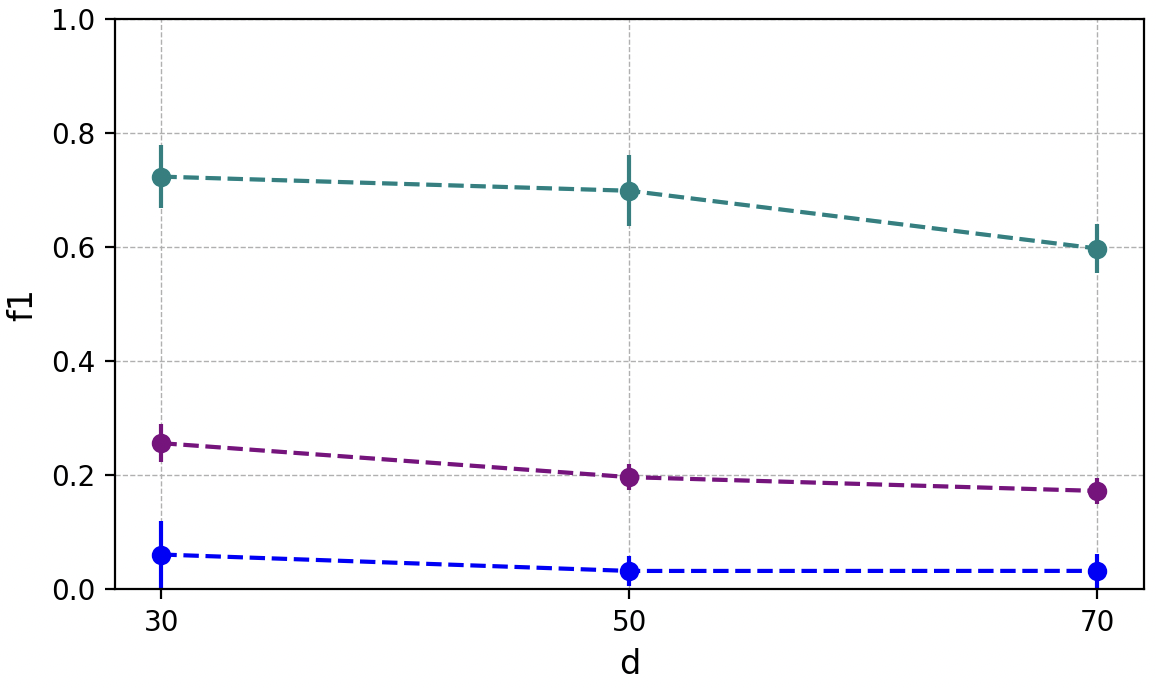}
    \end{subfigure}
    \begin{subfigure}[b]{0.32\textwidth}
        \includegraphics[width=\textwidth]%{figures/edgeprun3.png}
{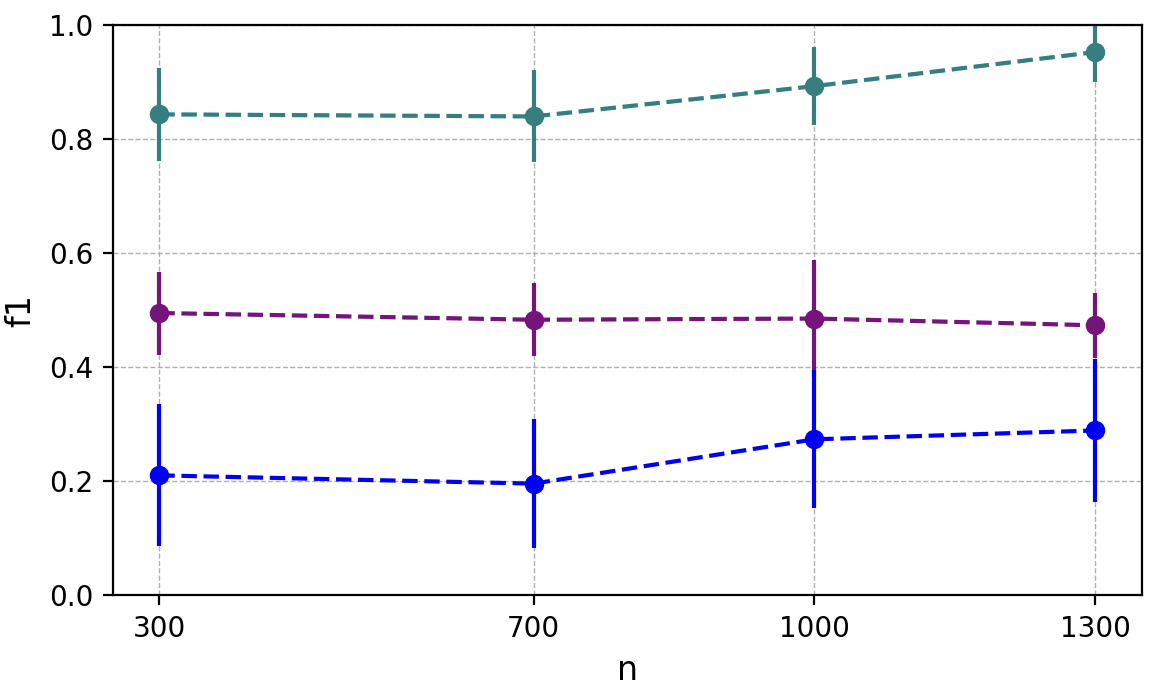}
    \end{subfigure}
    %\vspace{-5pt}
    \caption{Performance of ED on synthetic data, uniform noise. Left, middle: $n= 300$ with varying dimension $d$. Right: $d = 10$ with varying sample size $n$. %Data generated with (left) Gaussian, (middle) Uniform, and Laplacian (right) noise. 
    See Appendix \ref{appendix: edgepruneruntimes} for runtime results.} 
    \label{fig: edge_prune}
    %\vspace{-5pt}
\end{figure}

%\vspace{-3pt}
%We validate LHTS, NHTS and ED on synthetic data, demonstrating gains over previous methods.
\textbf{Setup$\;$} 
%In each experiment, we evaluate methods on 20 DAGs randomly generated with the Erdos-Renyi model \citep{erdos_renyi}. We set the average number of edges in each $d$-dimensional DAG as $d$. 
Methods\footnote{Code: \href{https://github.com/Sujai1/hybrid-discovery}{https://github.com/Sujai1/hybrid-discovery}.} are evaluated on 20 DAGs in each trial.  The DAGs are randomly generated with the
Erdos-Renyi model \citep{erdos_renyi}; the probability of an edge is set such that the average number of edges in
each $d$-dimensional DAG is $d$. Gaussian, Uniform, or Laplace noise is used as the exogenous error. In experiments for linear topological sorting methods (Figure \ref{fig: lin_sort}), we use linear causal mechanisms to generate the data; in experiments for nonlinear topological sorting methods (Figure \ref{fig: nonlin_sort}) and edge pruning algorithms (Figure \ref{fig: edge_prune}), we use quadratic causal mechanisms 
%nonlinear causal mechanisms 
to generate the data.  Existing ANM methods are prone to exploiting artifacts that are more common in simulated ANMs than real-world data \citep{reisach_beware_2021, reisach_scale-invariant_2023}, inflating their performance on synthetic DAGs and leaving real-world applicability an open question. To reduce concerns about such artifacts, data were generated with reduced $R^2$-sortability \citep{reisach_scale-invariant_2023}, and standardized to zero mean and unit variance \citep{reisach_beware_2021}.

%In each experiment, we run trials under two different conditions: in the leftmost graphs, we fix $n=500$ and vary the dimension $d$. In the rightmost graphs, we fix $d=10$, and vary the sample size $n$. In the first experiment (Figure \ref{fig: lin_sort}), we use linear causal mechanisms to generate the data, while in the second experiment (Figure \ref{fig: edge_prune}) we use quadratic causal mechanisms.

\textbf{Metrics$\;$} $A_{top} $ is equal to the percentage of edges that can be recovered by the returned topological ordering (an edge cannot be recovered if a child is sorted before a parent). We note that $A_{top}$ is a normalized version of the topological ordering divergence $D_{top}$ defined in \citep{rolland_score_2022}. Edge pruning algorithms return a list of predicted parent sets for each vertex: $F_1 = 2\frac{\text{Precision$\cdot$Recall}}{\text{Precision + Recall}}$ measures the performance of these predictions.

\textbf{Linear Topological Sorts$\;$}
%Ordering length refers to the length of the topological sort returned by the sorting algorithm: algorithms that produce linear sorts will be overlapping in the plots.
Figure~\ref{fig: lin_sort} demonstrates the performance of our linear topological ordering algorithm, LHTS, in comparison with the  benchmark algorithms,
% We take
DirectLiNGAM \citep{shimizu2011directlingam}  and $R^2$-Sort  \citep{reisach_scale-invariant_2023}.
%DirectLiNGAM is designed specifically for causal discovery in linear settings,  and 
$R^2$-Sort is a heuristic sorting algorithm that exploits artifacts common in simulated ANMs; both benchmarks are 
% as baseline comparators that are
agnostic to the noise distribution. 
We observe that both DirectLiNGAM and LHTS 
significantly outperform 
% the heuristic method 
$R^2$-sort. 
LHTS demonstrates asymptotic correctness in Figure~\ref{fig: lin_sort}(c), achieving near-perfect $A_{top}$ at $n=2000$. However, LHTS has consistently lower $A_{top}$ than DirectLiNGAM in Figure~\ref{fig: lin_sort}(a). On the other hand, LHTS encodes more causal information: the orderings produced by LHTS in Figure~\ref{fig: lin_sort}(b) had roughly $\sim70\%$ fewer layers than the orderings produced by DirectLiNGAM, reducing the size of potential parent sets $J$ by identifying many non-causal relationships. 
%This tradeoff between accuracy and encoded causal information is of significant interest: for high-dimensional applications with sparse causal graphs and ample data, such as genetic analysis, it may be worth it to slightly sacrifice sample complexity for a large reduction in 
% the size of 
%the causal ordering size. We aim to investigate the effects of this tradeoff more closely in future work. 

\textbf{Nonlinear Topological Sorts$\;$} Figure \ref{fig: nonlin_sort} illustrates the performance of our nonlinear topological sorting algorithm. We take GES \citep{chickering_learning_nodate}, GRaSP \citep{grasp_liam_2021}, GSP \citep{gsp_solus_2021}, DirectLiNGAM \citep{shimizu2011directlingam}, NoGAM \citep{montagna_causal_2023} , and $R^2$-Sort \citep{reisach_scale-invariant_2023} as baseline comparators that are all agnostic to the noise distribution. We excluded PC and RESIT since in general they perform much worse than baseline methods \citep{montagna_causal_2023}. We note that as DirectLiNGAM, NoGAM, and NHTS are FCM-based methods, they each return a unique topological ordering; however, as GES, GRaSP, and GSP are scoring-based methods \citep{glymour_review_2019}, they return only a MEC. All topological orderings contained within an MEC satisfy every conditional independence constraint in the data, and therefore are all \emph{equally valid}. To enable a fair comparison, we randomly select one ordering permitted by an outputted MEC for evaluation. We note that NHTS outperformed all baselines, achieving the highest median $A_{top}$ in all trials. Furthermore, as expected from Theorem \ref{theorem:nonlin_alg_time}, NHTS ran up to $4\times$ faster than NoGAM (see Appendix \ref{appendix: nonlin runtimes}). We provide additional experiments over DAGs with increasing edge density in Appendix \ref{appendix: nonlinear add exps}. 

%\vspace{-6pt}

\textbf{Edge Pruning$\;$}  Figure~\ref{fig: edge_prune} illustrates the performance of our edge pruning algorithm ED. We take covariate hypothesis testing with GAMs (CAM-pruning \citep{buhlmann_cam_2014}), Lasso regression, and RESIT as baseline comparators that are all agnostic to the noise distribution. All algorithms were given correct topological sorts: ED significantly outperformed all baselines, with the highest median $F_1$ score in all trials. ED was slower than Lasso, but was significantly faster than the other nonlinear edge pruning algorithms, CAM-pruning and RESIT. RESIT was excluded from higher-dimensional tests due to runtime issues.
The poor performance of baseline methods highlights the need for
a sample efficient nonparametric method
for accurate causal discovery of nonlinear DGPs. We provide additional experiments in settings with increasing density and varying noise distributions in Appendix \ref{appendix: pruing add exps}.

%\vspace{-6pt}

%\paragraph{Causal Discovery}
%Furthermore, as expected from Theorem \ref{theorem:nonlin_alg_time}, NHTS ran $8\times$ faster than NoGAM (see Appendix \ref{appendix: nonlin runtimes}).
%and NoGAM \citep{montagna_causal_2023} as baseline comparators for the full causal discovery pipeline.
%\Adjacency matrices were generated for random Erdos-Renyi graphs using the \texttt{igraph}\footnote{\href{https://r.igraph.org/}{https://r.igraph.org/}} R package, with edge directionality  imposed post-hoc by enforcing acyclicity. Using the \texttt{dowhy}\footnote{\href{https://www.pywhy.org/dowhy/}{https://www.pywhy.org/dowhy/}} Python package, data were sampled from ANMs with linear causal mechanisms and Laplace noise distributions, as defined by the random adjacency matrices. 
%\Existing ANM methods are prone to exploiting artifacts that are more common in simulated ANMs than real-world data \citep{reisach_beware_2021, reisach_scale-invariant_2023}, inflating their performance on synthetic DAGs and leaving real-world applicability an open question. To reduce concerns about such artifacts, our numerical evaluations use data with mean $R^2$-sortability less than or equal to $0.5$ across samples of the same size \citep{reisach_beware_2021, reisach_scale-invariant_2023}. $R^2$-sortability was computed using the \texttt{dodiscover}\footnote{\href{https://www.pywhy.org/dodiscover/}{https://www.pywhy.org/dodiscover/}} Python package.
% \section{Discussion}\label{sec: discussion}
\textbf{Discussion$\;$} In this paper we developed novel
global causal discovery algorithms by searching for and leveraging local causal relationships. We improved on previous topological ordering methods by running fewer regressions, each with lower dimensionality, producing hierarchical topological sorts. Additionally, we improved on previous edge pruning procedures by introducing a nonparametric constraint-based method that conditions on far fewer variables to achieve greater recovery of parent sets. We tested our methods on robustly generated synthetic data, and found that both our nonlinear sort NHTS and edge pruning algorithm ED significantly outperformed baselines.
Future work includes
extending the topological sorting algorithms to the full ANM setting with both linear and nonlinear functions, simultaneously exploiting both ancestor-descendent and parent-child relations, as well as adapting our approach to handle various forms of unmeasured confounding. Additionally, we aim to develop statistical guarantees of sample complexity for our methods, extending previous results \citep{sample_bound_2023} derived in the setting of nonlinear ANMs with Gaussian noise.

\newpage
\bibliography{references}
\raggedbottom
\newpage
\section*{Appendix}\label{sec:appendix}
\begin{appendices}
\section{Assumptions, Proofs, and Walkthrough for LHTS}
\subsection{Identifiability in Linear Setting}\label{appendix:lin_identify}
As an example \citep{peters_causal_2014}: suppose $X, N$ are normally distributed, $X \ind N$, and $Y = aX + N.$ If $\bar{a} = \frac{aVar(X)}{a^2Var(X)+\sigma^2}\neq \frac{1}{a}$ and $\bar{N}= X - \bar{a}Y$, then $\bar{N} \ind Y$ and the following DGP also fits the same distribution of $X,Y,N$: $X = \bar{a}Y + \bar{N}.$ To generalize the intuition that non-Gaussianity could lead to identifiability, \cite{shimizu_linear_2006}  use independent component analysis \citep{comon_ica_1994} to provide the following theorem: \begin{theorem}\label{thm:identify}
    Assume a linear SEM with graph $G$, where $x_i= \sum_{k(j)<k(i)}b_{ij}
   x_j
    + \varepsilon_i, \forall j = 1,\ldots,d,$ where all $\varepsilon_j$ are jointly independent and non-Gaussian distributed. Additionally, for each $j \in \{1,\ldots,d\}$ we require $b_{ij}\neq 0$ for all j $\in \textit{Pa}(x_i)$. Then, $G$ is identifiable from the distribution.
\end{theorem}

We assume that the identifiability conditions described in 
Theorem \ref{thm:identify} hold throughout Section \ref{sec:lin_setting}.
 
\subsection{Proof of Lemma \ref{lemma:causal_path_partition_ancestors}}\label{appendix:causal_path_partition_ancestors}
\causalPathPartition*
%kyra: restating the lemma here
\begin{proof}[Proof of Lemma~\ref{lemma:causal_path_partition_ancestors}]
For a pair of nodes $x_i$ and $x_j$, either $x_i\ind x_j$ or $x_i \nind x_j$. If the former, then $x_i,x_j$ are in AP1 relation. Suppose the latter is true: $x_i\nind x_j$ implies $\exists$ at least one active path between $x_i,x_j$. Active paths can either be backdoor or frontdoor paths: therefore either a frontdoor path exists, a backdoor path exists, or both a frontdoor and backdoor path exist. Thus, $x_i,x_j$ are either in AP2, AP3, or AP4 relation.
%kyra: this is okay. this is kind of trivial so might be able even able to get away without a proof.

\end{proof}
\subsection{Proof of Lemma \ref{lemma:path_ances_relation}}\label{appendix:path_ances_relation}
\LemmaPath*
\begin{proof}[Proof of Lemma~\ref{lemma:path_ances_relation}] Note, $x_i,x_j$ are ancestrally related if and only if there exists an active frontdoor path between them. The conclusion follows immediately.
%kyra: this is also okay    
\end{proof}
\subsection{Proof of Lemma \ref{lemma:AP1}}\label{appendix:AP1}
\APOne*
\begin{proof}[Proof of Lemma~\ref{lemma:AP1}] Note, $x_i \ind x_j$ if and only if there does not exist an active causal path between them. The conclusion follows immediately.
    %kyra: okay
\end{proof}
\subsection{Proof of Lemma \ref{lemma:AP2}}\label{appendix:AP2}
\APTwo*
\begin{proof}[Proof of Lemma~\ref{lemma:AP2}] 
Under an identifiable LiNGAM, and given a linear topological ordering $k: V\mapsto \mathbb{R}$, $x_i$ and $x_j$ admit the following representation:
% \kyra{need to explain constant and noise term, and note that the noise cannot follow gaussian.}
\begin{align*}
    x_i &= \sum_{k(m)<k(i)}\alpha_{im}\varepsilon_m + \varepsilon_i,
    \\
    x_j &= \sum_{k(m)<k(j)}\alpha_{jm}\varepsilon_m + \varepsilon_j, 
\end{align*}
where $\varepsilon_m$ are jointly independent noise terms and non-Gaussian distributed, following from Theorem~\ref{thm:identify}. 
Additionally, for each $m: k(m)<k(i)$ we require $\alpha_{im}\neq 0$ for all  $m \in \textit{Pa}(x_i)$. Similarly for $\alpha_{jm}$.

We first show the forward direction. Suppose $x_i,x_j$ are in AP2 relation.
Note, $x_i\nind x_j$. 
Let $\overline{M}$ be the complement of $M$, i.e., $\overline{M} = V\setminus M$. Then, $\exists$ the following decomposition of $x_i,x_j$:

\begin{align*}
    x_i &= \sum_{x_m\in M}\alpha_{im}\varepsilon_m +\sum_{x_m \in \overline{M}\cap \text{An}(x_i) }\alpha_{im}\varepsilon_m  +\varepsilon_i,
    \\
    x_j &= \sum_{x_m\in M}\alpha_{jm}\varepsilon_m +\sum_{x_m \in \overline{M}\cap \text{An}(x_j) }\alpha_{jm}\varepsilon_m  +\varepsilon_i.
\end{align*}
Then, the $x^M_i,x^M_j$ have the general form 
\begin{align*}
    x_i^M &= \sum_{x_m \in Y}\beta_{im}\varepsilon_m  +\varepsilon_i,
    \\
    x_j^M &= \sum_{x_m \in Z}\beta_{jm}\varepsilon_m  +\varepsilon_j,
\end{align*}
where $Y \subseteq \overline{M}\cap \text{An}(x_i)), Z \subseteq \overline{M}\cap \text{An}(x_j))$. Note, $Y \cap Z=\emptyset$. As $\varepsilon_i$ are all mutually independent, this implies that $r_i^j \ind r_j^i$.
\\

We now show the reverse direction. Suppose $x_i\nind x_j$, $r_i^j\ind x_i^M$, and $r_j^i\ind x_j^M$. Note that $x_i\nind x_j \implies$ $x_i,x_j$ are not in AP1 relation. Suppose for contradiction that $x_i,x_j$ are in AP3 or AP4 relation, where $x_i \in \text{An}(x_j)$ (WLOG). Then $\exists$ a frontdoor path between $x_i,x_j$, WLOG $x_i \in \text{An}(x_j)$. Regressing the descendent on the ancestor will always result in a dependent residual, even after projecting out the influence of the mutual ancestors. Therefore, $x^M_j \nind r_j^i$, leading to a contradiction. Therefore, $x_i,x_j$ must be in AP2 relation.
% kyra: this also looks fine
\end{proof}

\subsection{Proof of Lemma \ref{lemma:AP3}}\label{appendix:AP3}
\APThree*
\begin{proof}[Proof of Lemma~\ref{lemma:AP3}] We first show the forwards direction. Suppose $x_i,x_j$ are in $AP3$ relation. Note an active path exists between $x_i,x_j$, so $x_i \nind x_j$. WLOG, let $x_i \in \text{An}(x_j)$. Then, $x_i,x_j$ admit the following representation:
\begin{align*}
     x_i &= x_i
    \\
    x_j &= b_{ji}x_i + \sum_{x_m\in \text{An}(x_j)\setminus{x_i}}\alpha_{jm}x_m  +\varepsilon_j.
\end{align*}
Note, as there does not exist a backdoor path between $x_i,x_j$, we have $x_i \ind x_m \forall x_m \in \text{An}(x_j)\setminus{x_i}$. Therefore, $r_i^j \ind x_i$. Note that $x_j\in \text{De}(x_i)$: pairwise regression will leave $r_j^i \nind x_j$ as this an example of reverse causality, a special case of endogeneity \citep{pearl_causal_2016}.

We now show the reverse direction. WLOG, suppose $x_i,x_j$ satisfy condition 1) in Lemma \ref{lemma:AP3}. As $x_i \nind x_j$, they cannot be in AP1 relation. Suppose for contradiction that they are in AP2 or AP4 relation. Then, $\exists$ a backdoor path between $x_i,x_j$, and a confounding $x_z$ on that backdoor path. As $x_z$ confounds $x_i,x_j$ and is not adjusted for in a pairwise regression, we have $x_i \nind r_i^j, x_j \nind x_j^i$, a contradiction. Therefore, $x_i,x_j$ must be in AP2 relation.
\end{proof}
\subsection{Proof of Lemma \ref{lemma:AP4}}\label{appendix:AP4}
\APFour*
\begin{proof}[Proof of Lemma~\ref{lemma:AP4}]
    We first show the forward direction. Suppose $x_i,x_j$ are in AP4 relation. Note that an active path exists between $x_i,x_j$, so $x_i\nind x_j$. WLOG, suppose $x_i,\in\text{An}(x_j)$. Note, $x^M_i,x^M_j$ are the result of projecting mutual ancestors $M$ out of $x_i,x_j$. Therefore, $x^M_i,x^M_j$ admit the following representation:
    \begin{align*}
    x^M_i &= \sum_{x_m \in \overline{M}\cap \text{An}(x_i) }\alpha_{im}\varepsilon_m  +\varepsilon_i
    \\
    x^M_j &= \alpha_{ji}\varepsilon_i+\sum_{x_m \in \overline{M}\cap \text{An}(x_j)}\alpha_{jm}\varepsilon_m   +\varepsilon_j
\end{align*}
Note that $\overline{M}\cap \text{An}(x_i) \cap (\overline{M}\cap \text{An}(x_j)\cup x_j) = \emptyset$. Therefore, $r_i^j \ind x^M_i $. Note that $\varepsilon_i\in$ both $x^M_i,x^M_j$, therefore $r_j^i \nind x^M_j$.

We now show the reverse direction. Suppose condition 1) holds, i.e., $x_i \nind x_j$ and WLOG $r_i^j \ind x^M_i, r_j^i \nind x^M_j$. Suppose for contradiction that $x_i,x_j$ not in AP4 relation. They cannot be in AP1 relation because $x_i \nind x_j$. They cannot be in AP2 relation because $x^M_j \nind r_j^i$. By assumption they are not in AP3 relation. Therefore, we have a contradiction, therefore they are in AP4 relation.
\end{proof}

\subsection{Ancestor Sort}\label{appendix:ancestor_sort}
        \begin{algorithm}[H]
            \caption{\textbf{AS}: Ancestor Sort}
            \label{alg:as}
            \begin{algorithmic}[1]
            \State \textbf{input:} set of ancestral relations $\mathrm{ARS}$.
            \State \textbf{initialize:} hierarchical topological order $O$ [], remaining variables $\mathrm{RV}$ $[x_1,\ldots, x_d]$, layer $\mathrm{L}$ $[]$.
                \While{$\mathrm{RV}$ is not empty}
                \\\textbf{Stage 1: Determine Roots}
                    \For{$x_i\in RV$}
                        \If{$x_i$ has no ancestors $\in \mathrm{RV}$}
                            \State Append $x_i$ to $L$.
                        \EndIf
                    \EndFor
                \\\textbf{Stage 2: Update Sort and Remaining Variables}
                    \State Append $L$ to $O$.
                    \State Remove all vertices in $L$ from $RV$.
                \EndWhile
                \State \Return $O$
            \end{algorithmic}
        \end{algorithm}

\paragraph{Overview} In each iteration, this algorithm uses the set of ancestral relations to identify which vertices have no ancestors amongst the vertices that have yet to be sorted. It peels those nodes off, adding them as a layer to the hierarchical topological sort. First the roots are peeled off, then the next layer, and so on.

\paragraph{Proof of Correctness}
\begin{proof}
The input to the algorithm is an ancestor table $ARS$ cataloging all ancestral relations for every pair of nodes in an unknown DAG $G$. Let $H$ be the minimum number of layers ($H-1$ is the length of the longest directed path in the graph) necessary in a valid hierarchical topological sort of $G$.

\paragraph{Base Cases (1,2)}
\begin{enumerate}
    \item $H = 1$. None of the nodes have ancestors, so all nodes are added to layer 1 in Stage 1.
    \item $H = 2$. Nodes with no ancestors are added to layer 1 in Stage 1, and then removed from the remaining variables in Stage 2. As $H=2$, the longest directed path is 1, so all nodes in the remaining variables must have no ancestors in the remaining variables (otherwise, this would make the longest directed path 2). The nodes are added to layer 2 in Stage 1, and removed from the remaining variables in Stage 2. Remaining variables is empty, so now the order is correctly returned.
\end{enumerate}
\paragraph{Inductive Assumption} For any graph $G$ with $H < k$, Hierarchical Topological Sort will return a minimal hierarchical topological ordering.

We now show that hierarchical topological sort now yields a minimal hierarchical topological ordering for $G$ where $H=k$.
\begin{enumerate}
    \item  In the first iteration of Stage 1, nodes with no ancestors will be added to layer 1, then removed from the remaining variables in Stage 2.
    \item At this point, note that we can consider the induced subgraph formed by the nodes left in remaining, $G'$. The minimal hierarchical topological ordering that represents $G'$ must have  $k-1$ layers. It cannot be greater than $k-1$, because $G'$ is a subgraph of $G$ and we removed nodes in layer 1 of, reducing the maximal path length by 1. It cannot be less than $k-1$, because that would imply that $k$ was not the minimal number of layers needed to represent $G$.
    \item By Inductive Assumption, Hierarchical Topological sort will return the a correct minimal hierarchical topological ordering for the induced subgraph $G'$.
    \item Appending the layer 1 and the sort produced by $G'$ yields the full hierarchical topological sort for $G$.
\end{enumerate}

Inductive assumption is satisfied for $H=k$, so for a graph $G$ with arbitrary $H$, the algorithm recovers the correct hierarchical topological sort.

\end{proof}
\subsection{Proof of Correctness for Linear Hierarchical Topological Sort}
\begin{theorem}\ref{theorem:lin_alg}
Given a graph $G$, Algorithm \ref{algo:lin} asymptotically finds a correct hierarchical sort of $G$.
\end{theorem}
\begin{proof}\label{appendix:LHTS}
The goal is to show that all ancestral relations between distinct pair of nodes $x_i,x_j \in V$ determined by LHTS in Stages 1, 2, and 3 are correct. By \ref{appendix:ancestor_sort},  Stage 4 will return a valid hierarchical topological sort given fully identified ancestral relations.

Stage 1 identifies $x_i,x_j$ as in AP1 relaton if and only if $x_i \ind x_j$: it follows by Lemma \ref{lemma:AP1} that vertices in AP1 relation are correctly identified, and vertices in AP2, AP3 and AP4 relation are not identified.

Stage 2 identifies $x_i,x_j$ unidentified in Stage 1 as in AP3 relation if and only if after pairwise regression, one of the residuals is dependent, and the other is independent: it follows by Lemma \ref{lemma:AP3} vertices in AP3 relation are correctly identified, and vertices in AP2 or AP4 relation are not identified.

Consider a hierarchical topological sort of DAG $G$ $\pi_H$, with $h$ layers. Note, Layer 1 of $\pi_H$ consists of root nodes. Ancestral relations between root nodes and all other nodes are of type AP1 and AP3, and were discovered in Stage 1 and Stage 2.

We induct on layers to show that Stage 3 recovers all ancestral relations of type AP2 and AP4, one layer at a time in each iteration.

\paragraph{Base Iterations (1,2)}
\begin{enumerate}
    \item  All ancestral relationships are known for nodes in layer 1 of $\pi_H$, therefore all mutual ancestors of layer 2 and every lower layer are known. Then, by Lemma \ref{lemma:AP2} and Lemma \ref{lemma:AP4} ancestral relations of type AP2 and AP3 between vertices in layer 2 and all lower layers are discovered in iteration 1.
    \item All ancestral relationships are known for nodes in layer 1 and 2 of $\pi_H$, therefore all mutual ancestors of layer 3 and every lower layer are known. Then, by Lemma \ref{lemma:AP2} and Lemma \ref{lemma:AP4} ancestral relations of type AP2 and AP3 between vertices in layer 2 and all lower layers are discovered in iteration 2.
\end{enumerate}
\paragraph{Iteration $k-1$, Inductive Assumption} We have recovered all ancestral relationships of nodes in layers $0$ to $k-1$.
\begin{enumerate}
    \item As ancestral relationships of nodes in layers $0$ to $k-1$ are known, mutual ancestors of vertices in layer $k$ and every lower layer are known, so by Lemma \ref{lemma:AP2} and \ref{lemma:AP2}  all ancestral relations of type AP2 and AP4 between vertices in layer $k$ and every lower layer are discovered.
\end{enumerate}
Iteration $k-1$ Inductive Assumption is satisfied for iteration $k$, therefore we recover all ancestral relations of type AP2 and AP4 and 4) for nodes in layers $0$ to $k$. Thus, for a DAG with an arbitrary number of layers, Stage 3 recovers all ancestral relations of type AP2 and AP4.

\end{proof}

\subsection{Time Complexity Proof for Linear Hierarchical Topological Sort}\label{appendix:lhts_time}
\begin{proof}
    In Stage 1, LHTS runs $O(d^2)$ marginal independence tests that each have $O(n^2)$ complexity. In each step for Stage 2 and Stage 3, LHTS runs $O(d^2)$ marginal independence tests each with $O(n^2)$ complexity. In the worse case of a fully connected DAG, there are $d^2$ steps in total, across Stage 2 and Stage 3: this is because in each step one layer of the layered sort DAG is identified, and a fully connected DAG has $d$ layers. Therefore, the overall sample complexity of LHTS is $O(d^3n^2)$.
\end{proof}

\subsection{Walk-through of LHTS}\label{appendix:lhts_walk}
The following diagram illustrates each stage of LHTS on an exemplary 5-node DAG:

\begin{figure}[h]
    \centering
    \begin{subfigure}[b]{\textwidth}
        \includegraphics[width=\textwidth]{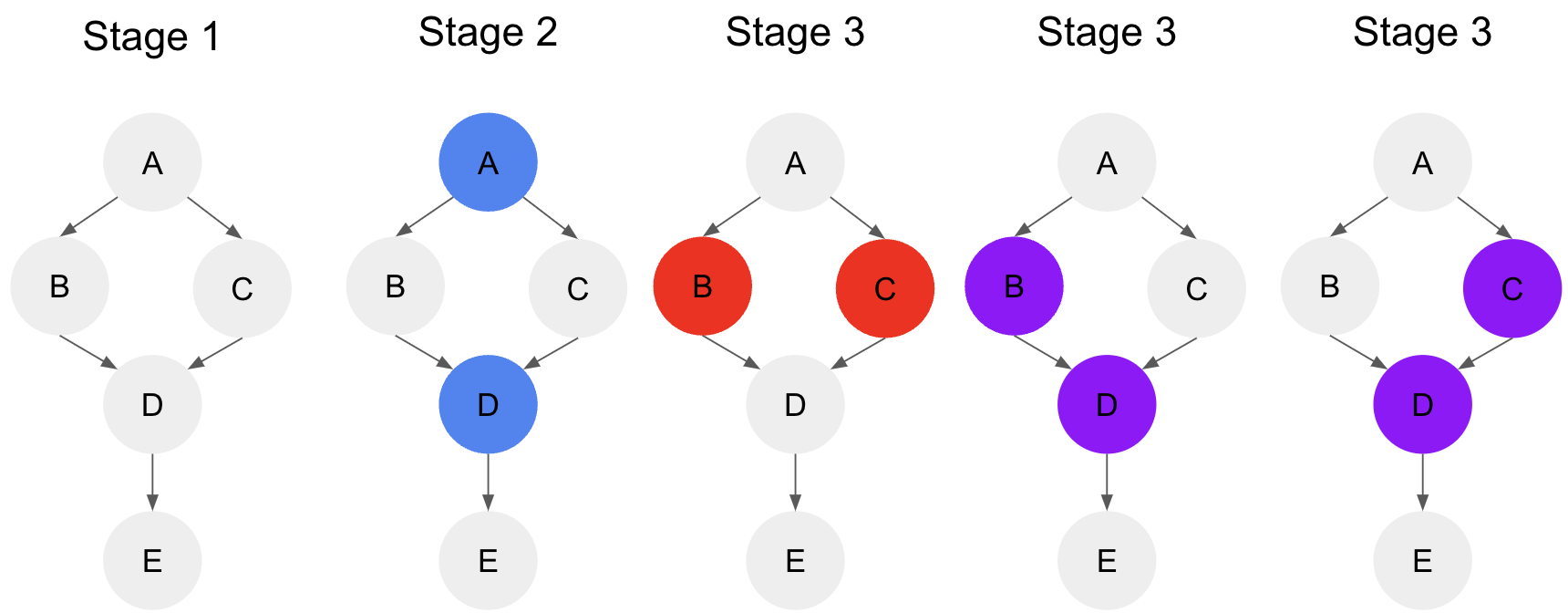}
    \end{subfigure}

\end{figure}

All vertices are dependent on each other, so no AP1 relations are discovered in Stage 1. In Stage 2, vertex A is discovered in AP2 relation to all vertices, and vertex D is discovered in AP2 relation to vertex E. In Stage 3, vertices B and C are discovered to be in AP3 relation to each other; then, we discover vertex B in AP4 relation to vertex D, and vertex C in AP4 relation to vertex D. Therefore, in Stage 4, we recover the topological sort after running the AS subroutine.

\subsection{Proof of Lemma~\ref{lemma:root_vertex}: Root Active Causal Ancestral Path Relations}\label{lemma:roots}
\rootVertex*
% A vertex $x_i$ is a root if and only if $x_i$ is in either AP1 or AP3 (as an ancestor) relations with all other nodes.
\begin{proof}[Proof of Lemma~\ref{lemma:root_vertex}]
We first show the forward direction. If $x_i$ is a root, then $\text{De}(x_i)= \emptyset$. Therefore, it cannot have a backdoor path between it and any other vertex, therefore it must be in either $AP1$ or $AP3$ relation with other vertices.

We now show the reverse direction. If $x_i$ is in AP1 or AP3 (as an ancestor) relations with all other nodes, it cannot have any parents. Therefore, $x_i$ is a root.

\end{proof}
\end{appendices}
\newpage
\begin{appendices}
\section{Assumptions, Proofs, and Walkthrough for NHTS}
\subsection{Identifiability in Nonlinear Setting}\label{appendix:nonlin_identify}
Following the style of \citep{montagna_assumption_2023}, we first observe that the following condition guarantees that the observed distribution of a pair of variables $x_i,x_j$ can only be generated by a unique ANM:
\begin{condition}[\citet{hoyer_bayesian_2009}]\label{theorem:bivariate_anm}
Given a bivariate model $x_i = \varepsilon_i, x_j = f_j(x_i)+\varepsilon_j$ generated according to \eqref{eq:ANM}, we call the SEM an identifiable bivariate ANM  if the triple $(f_i,p_{\varepsilon_i},p_{\varepsilon_j})$ does not solve the differential equation $k'''=k''(-\frac{g'''f'}{g''}+\frac{f''}{f'})-2g''f''f'+g'f'''+\frac{g'g'''f''f'}{g''}-\frac{g'(f'')^2}{f'}$ for all $x_i,x_j$ such that $f'(x_i)g''(x_j-f_j(x_i))\neq0$, where $p_{\varepsilon_i},p_{\varepsilon_j}$ are the density of $\varepsilon_i,\varepsilon_j$, $f=f_j,k=\log p_{\varepsilon_i}, g=p_{\varepsilon_j}$. The arguments $x_j-f_j(x_i),x_i$ and $x_i$ of $g,k$ and $f$ respectively, are removed for readability.
\end{condition}

There is a generalization of this condition to the multivariate nonlinear ANM proved by \citep{peters_causal_2014}:
\begin{theorem}\label{theorem:multivariate_anm}
(Peters et al. \citep{peters_causal_2014}). An ANM corresponding to DAG $G$ is identifiable if $\forall x_j \in V, x_i \in \text{Pa}(x_j)$ and all sets $S \subseteq V$ with $\text{Pa}(x_j) \setminus \{i\} \subseteq S \subseteq \overline{\text{De}(j)}\setminus{\{x_i, x_j\}}$, $\exists$ $X_S$ with positive joint density such that the triple $\Big{(}f_j(\text{Pa}(j)\setminus{\{x_i\}},x_i), p_{x_i|X_s},p_{\varepsilon_j}\Big{)}$ satisfies Condition \ref{theorem:bivariate_anm}, and $f_j$ are non-constant in all arguments.
\end{theorem}

We assume that the identifiability conditions described in 
Theorem \ref{theorem:multivariate_anm} hold throughout Section \ref{sec:nonlin_setting}.

%We also assume that, beyond causal sufficiency, if $\exists$ an unmeasured mediator $x_k$ between observed vertices $x_i, x_j \in V$, then $x_k$ is a linear function of $x_i$, and $x_j$ is a linear function of $x_k$.
%We also assume that, beyond causal sufficiency, for any observed vertex $x_i$, there does not exist any unobserved causes of $x_i$.

%if $\exists$ an unmeasured mediator $x_k$ between observed vertices $x_i, x_j \in V$, then $x_k$ is a linear function of $x_i$, and $x_j$ is a linear function of $x_k$.

\subsection{Proof of Lemma \ref{lemma:causal_parent_path_partition}}\label{appendix:causal_parent_path_partition}

\parentenum*

\begin{proof}
    Either $x_i\ind x_j$ or $x_i \nind x_j$: if the former, then $x_i,x_j$ are in PP1 relation. Suppose the latter is true. Then, either $x_i\ind C$ or $x_i \nind C$: if the former, then $x_i,x_j$ are in PP2 relation. Suppose the latter is true. Then, either $x_i \in \text{PA}(x_j)$ or $x_i \not\in \text{PA}(x_j)$. If the former, then $x_i,x_j$ are in PP3 relation, and if the latter, then $x_i,x_j$ are in PP4 relation.
\end{proof}

\subsection{Proof of Lemma \ref{lemma:PP1}}\label{appendix:PP1}
\nonppone*
\begin{proof}
 We first show the forward direction. If $x_i,x_j$ are not in PP1 relation, then either $x_i \in \text{Pa}(x_j)$ or there exists an active causal path between $x_i,x_j$; therefore, $x_i \nind x_j$.

We now show the reverse direction. If $x_i\nind x_j$ there exists an active causal path between $x_i,x_j$, therefore they cannot be in PP1 relation.
\end{proof}
\subsection{Proof of Lemma \ref{lemma:PP2}}\label{appendix:PP2}
\pptwo*
\begin{proof}
Note, there exists two sub cases of PP2 relation: PP2a) $x_i$ is the only parent of $x_j$, i.e. $|C|=0$ and PP2b) $x_i$ is not the only parent of $x_j$, i.e., $|C|>0$.

We first show the forward direction. Suppose $x_i,x_j$ are in PP2a relation. Then, as the 
\begin{align*}
    x_j = f_j(x_i) +\varepsilon_j
\end{align*}
$x_i \ind r^j_i$.
Suppose that $x_i, x_j$ are in a PP2b relation. Consider $P_{ij}$: note that $x_i \ind C$, and $x_j \nind C$, $C \subseteq P_{ij}$. Note that $\text{De}(x_j) \nind x_i$, therefore $\text{De}(x_j) \not \in P_{ij}$. Therefore, $P_{ij}$ contains all parents of $x_j$, and excludes all descendent of $x_j$.
Then, as 
\begin{align*}
    x_j = f_j(x_i,C) +\varepsilon_j
\end{align*}
we have $x_i \ind r^i_{j_P}$.

We now show the reverse direction. Suppose $x_i \ind r_i^j$. Note by assumption $x_i,x_j$ are not in PP1 relation. Suppose for contradiction that $x_i,x_j$ are in PP3 or PP4 relation. Then, there must exist at least one active path between $x_i,x_j$ that goes through $C$. If $\exists$ a backdoor path from $x_i,x_j$, then at least one vertex in $C$ is a confounder of $x_i,x_j$, and thus $x_i \nind r_i^j$; therefore, there must exist a frontdoor path between $x_i,x_j$ through $C$, with at least one mediator $x_k$. Note that $x_k$ is dependent on both $x_i$ and $x_j$, and therefore is excluded from $P_{ij}$. The exclusion of $x_k$ introduces omitted variable bias \citep{pearl_causal_2016}, and thus $x_i \nind r^j_{i_P}$, leading to contradiction. Thus, $x_i,x_j$ are in PP2 relation.

\end{proof}
\subsection{Proof of Lemma \ref{lemma:root_identification_nonlinear}}\label{appendix:root_id_nonlinear}
\rootlemma*
\begin{proof}
We first show that all non-isolated root vertices are contained in $W$. Note, if a root $x_i$ is not isolated, it has at least one child $x_j$. By definition, $x_i$ has no parents, so there cannot exist a backdoor path between $x_i,x_j$. Consider $x_j \in \text{Ch}(x_i)$ that is in a hierarchical layer closest to $x_j$.
 Note that there cannot exist an active path from $x_i$ to $\text{Pa}(x_j)\setminus{x_i}$, because that would imply that $x_j$ is not the child of $x_i$ that is in the closest hierarchical layer.
Therefore, $x_i,x_j$ must be in PP2 relation, and $x_i \in W$.

We now show the forward direction. Suppose $x_i$ is a root vertex. Then, $x_i$ has no parents, so it cannot be the descendent of any vertex, let alone any vertex in $W$: Condition 1) is satisfied. For each $x_j\in W$, there either exists an active path between $x_i,x_j$, or there does not. If there does not exist an active path,
% then $x_i \ind x_j$ 
then $x_i \ind x_j$, satisfying Condition a). 
If there does exist an active path, then as $x_i$ is a root, $x_j\in \text{De}(x_i)$: the active path must be a frontdoor path. Consider the vertex $x_k \in \text{Ch}(x_j)$ that is in a hierarchical layer closest to $x_j$, AND is along the active frontdoor path from $x_i$ to $x_j$. Note that $x_i$ is in PP2 relation with $x_k$.
If $x_k = x_j$, then Condition b) is satisfied. If $x_k \neq x_j$, then we note that there is an active frontdoor path between $x_k$ and $x_j$. Therefore, $x_k,x_j$ are dependent even after conditioning on the mutual ancestor $x_i$, i.e., $x_j \nind x_k|x_i$: Condition c) is satisfied. Therefore, Condition 2) is always satisfied.

%insert visualization here

We now show the reverse direction. Suppose Condition 1) and Condition 2) are satisfied for $x_j \in W$. Suppose for contradiction that $x_j$ is not a root. Note that $\exists x_i \in W$ such that $x_i$ is a root and $x_j \in \text{De}(x_i)$. Consider any vertex $x_k \in \text{Ch}(x_j)$ such that $x_j$ is in PP2 relation to $x_k$. Note that all active paths between $x_i$ and $x_k$ must be frontdoor paths, and $x_j$ must lie along all these active frontdoor paths: otherwise, this would contradict that $x_j$ and $x_k$ are in PP2 relation. Therefore, $x_i \ind x_k |x_j$. This contradicts Condition 2c), therefore Condition 1) and Condition 2) cannot hold true for $x_i \in W$ that is not a root.

%insert visualization here
    
\end{proof}
\subsection{Proof of Correctness for Nonlinear Hierarchical Topological Sort}\label{appendix:NLHTS_proof}

\algotwocorrect*
\begin{proof}
This proof can be broken into two parts: we first show 1) all root vertices are identified after Stage 3, then we show 2) that Stage 4 recovers a correct hierarchical topological sort given the root vertices.

Note that Stage 1 identifies $x_i,x_j$ as not in PP1 relation if and only if $x_i \nind x_j$: it follows by Lemma \ref{lemma:PP1} that these identifications are correct.

Note that Stage 2 identifies the set $x_i\in W$ if and only if $\exists x_j$ such that either 1) $x_i \ind r_i^j$ or 2) $x_i \ind r^j_{i,P}$: it follows by Lemma \ref{lemma:PP2} that these identifications are correct.

Note that Stage 3 explicitly uses a condition from Lemma \ref{lemma:root_identification_nonlinear} to find all non-isolated root vertices - it is therefore correct. Note that all isolated root vertices were identified in Stage 1. Therefore, all roots are identified.

Consider a hierarchical sort of DAG $G$ $\pi_L$: suppose the maximal directed path in $G$ has size $h$. Note that layer 0 of $\pi_L$ consists of root nodes, and therefore has been identified. We note the following \textbf{observation}: when $x_i$ is nonparametrically regressed on all $x_k \in \pi_L$ to obtain residual $r_i$, if $\exists$ an ancestor of $x_i$ not in $\pi_L$, then $r_i$ will be dependent on at least one $x_i \in \pi_L:$ due to omitted variable bias \citep{pearl_causal_2016}.

We now induct on the layers of $\pi_L$ to show that Stage 4 correctly recovers all layers.

\paragraph{Base Iterations (1,2)}
\begin{enumerate}
    \item  All roots are identified, so layer 0 of $\pi_L$ is identified.
    \item By \textbf{observation}, only $x_i$ in layer 1 of $\pi_L$ will have independent residuals $r_i$ after nonparametric regression on $x_i \in \pi_L$, so layer 1 is correctly identified.
\end{enumerate}
\paragraph{Iteration $k-1$, Inductive Assumption} We have recovered all layers of $\pi_L$ up to layer $k-1$.
\begin{enumerate}
\item By \textbf{observation}, only $x_i$ in layer $k$ of $\pi_L$ will have independent residuals $r_i$ after nonparametric regression on $x_i \in \pi_L$, so layer $k$ is correctly identified.
\end{enumerate}
Iteration $k-1$ Inductive Assumption is satisfied for iteration $k$, therefore we recover all layers of $\pi_L$ from $0$ to $k$. Thus, for a DAG with an arbitrary number of layers, Stage 4 recovers the full hierarchical sort.
\end{proof}

\subsection{Time Complexity for NHTS}\label{appendix:nhts_time}
\algotworun*
\begin{proof}
    In Stage 1, NHTS runs $O(d^2)$ marginal independence tests that each have $O(n^3)$ complexity. In Stage 2, NHTS runs $O(d^2)$ nonparametric regressions and $O(d^2)$ marginal independence tests, each of which have $O(n^3)$ complexity. In Stage 3 NHTS runs at most $O(d^2)$ conditional independence tests, each of which has $O(n^3)$ complexity. In the worst case of a fully connected DAG, NHTS goes through $O(d)$ steps in Stage 4: in each step of Stage 4, NHTS runs $O(d)$ nonparametric regressions and $O(d^2)$ marginal independence tests, each of which has $O(n^3)$ complexity. Therefore, the overall sample complexity of NHTS is $O(d^3n^3)$.
\end{proof}
\subsection{Walk-through of NHTS}\label{appendix:nhts_walk}
The following diagram illustrates each stage of NHTS on an exemplary 5-node DAG:

\begin{figure}[h]
    \centering
    \begin{subfigure}[b]{\textwidth}
        \includegraphics[width=\textwidth]{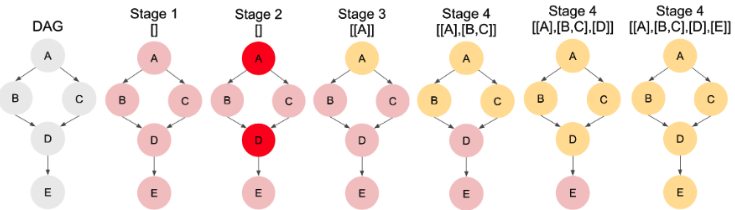}%{figures/nhts_walkthrough_2.png}
    \end{subfigure}

\end{figure}

In Stage 1, we discover that none of the vertices are in PP1 relations. In Stage 2, we discover that vertex A is in PP2 relation to vertices B and C, and vertex D is in PP2 relation to vertex E: therefore, A and D are our potential roots. In Stage 3, we find that vertex D d-separates its child, vertex E, from vertex A: therefore A is the root vertex. In Stage 4, we regress the unsorted vertices onto A, finding that B,C are independent of the residual in the first round, then D, then E in the last round.

\subsection{Time Complexity of NHTS vs RESIT, NoGAM}\label{appendix:nhts_time_red}
\nhtsrunred*
\begin{proof}
    
RESIT and NoGAM both identify leaf vertices in an iterative fashion, regressing each unsorted vertex on the rest of the unsorted vertices; RESIT tests the residual for independence with the covariate set while NoGAM uses the residual for score matching. Therefore, the number of regressions run in both methods in each step equals
one plus the covariate set size. Therefore, when the covariate set size is $k > \frac{d}{2}$, there are $k+1$ regressions run. 
\\~\\
In the case of a fully directed graph, the first stage of NHTS only runs pairwise regressions with empty conditioning sets. After the first stage, NHTS regresses each unsorted vertex onto all sorted vertices, finding vertices with independent residuals. Therefore, the number of regressions run is equal to $d$ minus the size of the covariate set. Therefore, when the covariate set size is $k > \frac{d}{2}$, there are $d - k$ regressions run.
\end{proof}

\end{appendices}

\newpage
\begin{appendices}
\section{Proofs and Walkthrough for ED}
\subsection{Proof of Lemma}\label{appendix:edge_lemma}
\parentedge*
\begin{proof}
     Note, $C_{i,j}$ blocks all backdoor paths between $x_i, x_j$ and $M_{i,j}$ blocks all frontdoor paths between $x_i,x_j$, except the direct edge.
    
    We first show the forward direction. If $x_i \not\ind x_j |Z_{i,j}$, there must be an direct edge between $x_i,x_j$, and as $k(i)<k(j)$, we have $x_i\rightarrow x_j$.

    We now show the reverse direction. If $x_i \rightarrow x_j$, then there does not exist a conditioning set that makes $x_i \ind x_j$, which implies $x_i \not \ind x_j |Z_{i,j}$.
\end{proof}
\subsection{Hierarchical Version of Edge Discovery}\label{appendix:hierarchical_edge}
The current implementation of Edge Discovery takes a linear hierarchical topological sort as input; this is equivalent to a hierarchical topological sort where every layer contains only one vertex. To generalize ED to a hierarchical sort, we simply adjust how the algorithm loops over the sort, and which vertices are included in which conditioning sets $C_{ij},M_{ij}$. We give an example of how the latter would change: suppose the algorithm is checking whether $x_i\in\text{Pa}(x_j)$ where $x_i$ is in layer 2 ($\pi_L(x_i)=2$) and $x_j$ is layer 6 ($\pi_L(x_j)=6$). $M_{ij}$ would be equal to the vertices who are parents of $x_j$ that are in layers that are between layer 2 and 6: i.e. $M_{ij}$ equals $x_k \in \text{Pa}(x_j)$ such that $2<\pi_L(x_k)<6$. $C_{ij}$, would stay the same, just being equal to $\text{Pa}(x_i)$. Now, to generalize the looping part, notice that the linear version of ED loops over indices of the linear topological sort. The hierarchical version of Edge Discovery would loop over and within layers of the topological sort. We give an example of this: suppose the algorithm has found all parents of vertices in layer 4 or lower: the next step would be to find the parents of vertices in layer 5. The algorithm sets any vertex in layer 5 as $x_j$, and first finds all parents of $x_j$ in the immediately preceding layer, layer 4. The algorithm then finds all parents of $x_j$ in the layer preceding layer 4, layer 3. This is essentially the same approach as the linear version, except the size of the conditioning sets and number of conditional tests run is far lower, as the layers provide extra knowledge, limiting which variables can be parents or confounders of other variables. 

\subsection{Edge Discovery}\label{appendix:edge_discovery}
\begin{figure}[h!]
    \centering
        \includegraphics[width=\textwidth]{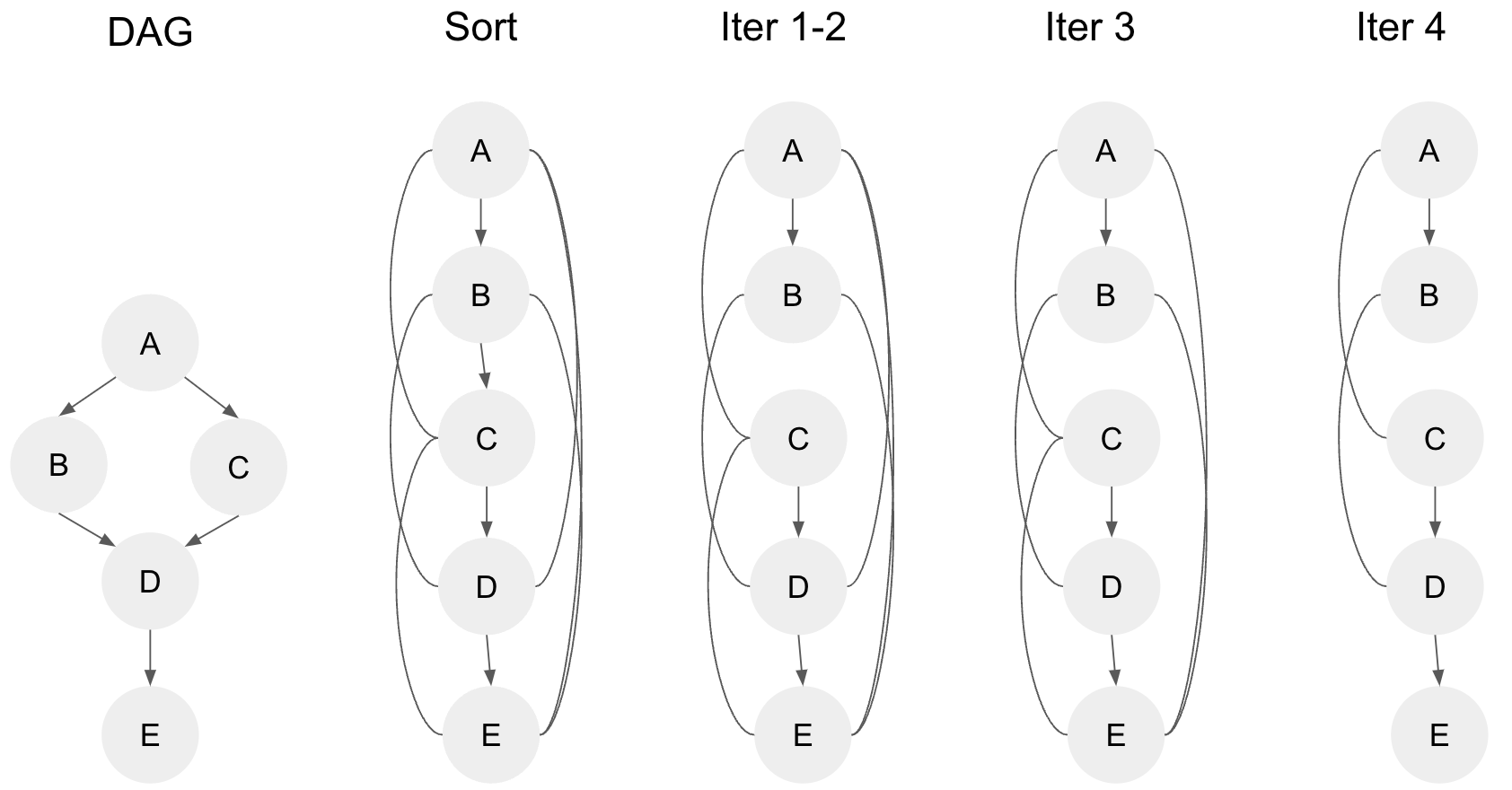}
\end{figure}
We first provide a walk-through of ED on an exemplary 5-node DAG:

We first use a topological ordering algorithm to obtain a sort from the DAG: the sort corresponds to a fully connected DAG, from which we will prune spurious edges. In the first two iterations we find that $A \rightarrow B$, but $B \not\rightarrow C$, as $B \ind C | A$. In the third iteration we find that $B\rightarrow D, C\rightarrow D$, but $A\not\rightarrow D$, as $A \ind D | B,C$. In the final iteration, we find that $D\rightarrow E$, but $A\not\rightarrow E, B \not\rightarrow E, C\not\rightarrow E$, given $A \ind E |D, B \ind E| D, C \ind E|D$. Therefore, we recover the correct set causal edges, removing all spurious edges.
We now present a statement of correctness and proof for ED:
\edcorrect*
\paragraph{Proof of Correctness}
\paragraph{Background}
We are given a linear topological sort $\pi(V)= [x_1,x_2,\ldots,x_d]$ (where $x_1$ has no parents). For $x_a$ that appears before $x_b$, let $Z_{a,b} = C_{a,b} \cup M_{a,b}$, where $C_{a,b}$ is the set of potential confounders of $x_a,x_b$, and $M_{a,b}$ is the set of known mediators of $x_a,x_b$ (defined precisely in Lemma \ref{lemma:edge}). Let $x \ind y|\cdot$ be the value of a conditional independence test between $x$ and $y$.
\begin{proof}~
\paragraph{Base Case Iterations (1,2,3,4)} 
\begin{enumerate}
    \item Finding parents of $x_1$: by the topological sort, $x_1$ has no parents.
    \item Finding parents of $x_2$: 
    As $x_1$ is the only possible parent of $x_2$, there are no possible confounders or mediators between $x_1$ and $x_2$, so we initialize $Z_{1,2}=\emptyset$, then
    by $L3$ $x_1\not\ind x_2 |Z_{1,2} = x_1 \not\ind x_2 \iff x_1\rightarrow x_2$. Thus, we recover all possible edges between $[x_1,x_2]$.
    
    \item Finding parents of $x_3$: first we check whether $x_2\rightarrow x_3$, then we check whether $x_1\rightarrow x_3$. From iteration 2, we know all edges between $[x_1,x_2]$.
    
    \textbf{Case 1}: if $x_1\rightarrow x_2$, then it is a possible confounder. 
    Therefore, we initialize  $C_{2,3}=x_1$. By topological sort there is no mediator between $x_2$ and $x_3$, therefore we initialize $M_{2,3}=\emptyset$.
     Then by Lemma \ref{lemma:edge} $x_2\not\ind x_3|Z_{2,3} = x_2 \not\ind x_3 | x_1$ $\iff $ $x_2\rightarrow x_3$. If an edge between $x_2$ and $x_3$ exists, initialize ${M_{1,3}}$ accordingly. Note, $x_1$ has no parents, so we initialize $C_{1,3}=\emptyset$.  Then, by Lemma \ref{lemma:edge} $x_1\not\ind x_3|Z_{1,3}\iff$ $x_1\rightarrow x_3$. We recover all possible edges between $[x_1,x_2,x_3]$.   
     
    \textbf{Case 2}: if $x_1\not\rightarrow x_2$, there are no possible confounders between $x_2$ and $x_3$. Therefore, we initialize $C_{2,3}=\emptyset$.  By topological sort there is no mediator between $x_2$ and $x_3$, therefore we initialize $M_{2,3}=\emptyset$. Then by Lemma \ref{lemma:edge} $x_2\not\ind x_3|Z_{2,3} = x_2 \not\ind x_3$ $\iff $ $x_2\rightarrow x_3$. Note, $x_1$ has no parents, so we initialize $C_{1,3}=\emptyset$. Note, as $x_1\not\rightarrow x_2$, there are no possible mediators between $x_1$ and $x_3$: we initialize $M_{1,3}=\emptyset$. Then, by Lemma \ref{lemma:edge} $x_1\not\ind x_3|Z_{1,3}=x_1\not\ind x_3$ $\iff$ $x_1\rightarrow x_3$. We recover all possible edges between $[x_1,x_2,x_3]$.
    
    \item Finding parents of $x_4$: first we check whether $x_3\rightarrow x_4$, then whether $x_2\rightarrow x_4$, then whether $x_1\rightarrow x_4$. From iteration 3, we know all edges between $[x_1,x_2,x_3]$.
    
    \textbf{Case 1}: if $[x_1,x_2,x_3]$ has no edges, then no node causes $x_3$ directly or indirectly, therefore we initialize $C_{3,4}=\emptyset$.
    There are no possible mediators between $x_3$ and $x_4$, so $M_{3,4}=\emptyset$. Therefore, by Lemma \ref{lemma:edge} $x_3\not\ind x_4|Z_{3,4}=x_3\not
    \ind x_4$$\iff$ $x_3 \rightarrow x_4$. As $[x_1,x_2,x_3]$ has no edges, no node causes $x_2$ directly or indirectly, therefore we initialize $C_{2,4}=\emptyset$. Note, $x_2\not\rightarrow x_3$, so there are no possible mediators between $x_2$ and $x_4$, so we initialize $M_{2,4}=\emptyset$. Then, by Lemma \ref{lemma:edge} $x_2\not\ind x_4|Z_{2,4}=x_2\not\ind x_4$$\iff$ $x_2\rightarrow x_4$. Note, $x_1$ has no parents, so we initialize $C_{1,4}=\emptyset$. As, $x_1$ does not cause $x_2$ or $x_3$ there are no possible mediators between $x_1$ and $x_4$, therefore we initialize $M_{1,4}=\emptyset$. Then, by Lemma \ref{lemma:edge} $x_1\not\ind x_4|Z_{1,4}=x_1\not\ind x_4 \iff x_1\rightarrow x_4$. We recover all possible edges between $[x_1,x_2,x_3,x_4]$.
    
    \textbf{Case 2}: if $x_1\rightarrow x_2$ is the only edge between the nodes $[x_1,x_2,x_3]$, then no node causes $x_3$ directly or indirectly, therefore $C_{3,4}=\emptyset$. By topological sort there is no mediator between $x_3,x_4$, so we initialize $M_{3,4}=\emptyset$. Then, by Lemma \ref{lemma:edge} $x_3\not\ind x_4|Z_{3,4}=x_3\not\ind x_4\iff x_3\rightarrow x_4$. As $x_1\rightarrow x_2$, we initialize $C_{2,4}=x_1$. As $x_2\not\rightarrow x_3$, there are no possible mediators between $x_2$ and $x_4$, so we initialize $M_{2,4}=\emptyset$. Then, by Lemma \ref{lemma:edge} $x_2\not\ind x_4|Z_{2,4}= x_2\not\ind x_4|x_1 \iff x_2\rightarrow x_4$. If an edge exists between $x_2$ and $x_4$, we initialize $M_{1,4}$ accordingly. As $x_1$ has no parents, we initialize $C_{1,4}=\emptyset$. Then, by Lemma \ref{lemma:edge} $x_1\not\ind x_4|Z_{1,4}\iff x_1 \rightarrow x_4$. We recover all possible edges between $[x_1,x_2,x_3,x_4]$.
    
    \textbf{Case 3}: if $x_1\rightarrow x_3$ is the only edge between the nodes $[x_1,x_2,x_3]$, then $x_1$ is the only potential confounder of $x_3$ and $x_4$ so we initialize $C_{3,4}=x_1$. There are no possible mediators between $x_3$ and $x_4$, so we initialize $M_{3,4}=\emptyset$. Then, by Lemma \ref{lemma:edge} $x_3\not\ind x_4 |Z_{3,4}=x_3\not\ind x_4|x_1 \iff x_3\rightarrow x_4$. Note, $x_2$ has no parents, so we initialize $C_{2,4}=\emptyset$. Note, $x_2\not\rightarrow x_3$, so we initialize $M_{2,4}=\emptyset$. Then, by Lemma \ref{lemma:edge} $x_2\not\ind x_4 |Z_{2,4}=x_2\not\ind x_4 \iff x_2\rightarrow x_4$. If an edge exists between $x_3$ and $x_4$, we initialize $M_{1,4}$ accordingly. As $x_1$ has no parents, we initialize $C_{1,4}=\emptyset$. Then, by Lemma \ref{lemma:edge} $x_1\not\ind x_4|Z_{1,4} \iff x_1 \rightarrow x_4$. We recover all possible edges between $[x_1,x_2,x_3,x_4]$.
    
    \textbf{Case 4}: if $x_1\rightarrow x_2,x_2\rightarrow x_3$ are the only edges between nodes $[x_1,x_2,x_3]$, then $x_1$ and $x_2$ cause $x_3$ either indirectly or directly. Therefore, we initialize $C_{3,4}=\{x_1,x_2\}$. There are no possible mediators between $x_3$ and $x_4$, so we initialize $M_{3,4}=\emptyset$. Then, by Lemma \ref{lemma:edge} $x_3\not\ind x_4|Z_{3,4}=x_3\not\ind x_4|x_1,x_2 \iff x_3 \rightarrow x_4$. If an edge exists between $x_3$ and $x_4$, we initialize $M_{2,4}$ accordingly. Note, $x_1$ is a parent of $x_2$, so we initialize $C_{2,4}=x_1$. Then, by Lemma \ref{lemma:edge} $x_2 \not \ind x_4 |Z_{2,4}\iff x_2\rightarrow x_4$. If an edge exists between $x_3$ and $x_4$, and/or between $x_2$ and $x_4$, initialize $M_{1,4}$ accordingly. Note, $x_1$ has no parents, so initialize $C_{1,4}=\emptyset$. Then by Lemma \ref{lemma:edge} $x_1\not\ind x_4|Z_{1,4}\iff x_1\rightarrow x_4$. We recover all possible edges between $[x_1,x_2,x_3,x_4]$.

    \textbf{Case 5}: if $x_1\rightarrow x_3,x_2\rightarrow x_3$are the only edges between nodes $[x_1,x_2,x_3]$, then $x_1$ and $x_2$ cause $x_3$ directly. Therefore, we initialize $C_{3,4}=\{x_1,x_2\}$. There are no possible mediators between $x_3$ and $x_4$, so we initialize $M_{3,4}=\emptyset$. Then, by $L3$ $x_3\not\ind x_4|Z_{3,4}=x_3\not\ind x_4|x_1,x_2 \iff x_3 \rightarrow x_4$. If an edge exists between $x_3$ and $x_4$, we initialize $M_{2,4}$ accordingly. Note, $x_2$ has no parents, so we initialize $C_{2,4}=\emptyset$. Then,by $L3$ $x_2 \not \ind x_4 |Z_{2,4}\iff x_2\rightarrow x_4$. If an edge exists between $x_3$ and $x_4$, initialize $M_{1,4}$ accordingly. Note, $x_1$ has no parents, so initialize $C_{1,4}=\emptyset$. Then by $L3$ $x_1\not\ind x_4|Z_{1,4}\iff x_1\rightarrow x_4$. We recover all possible edges between $[x_1,x_2,x_3,x_4]$.
    
    \textbf{Case 6}: if $x_1\rightarrow x_2,x_1\rightarrow x_3, x_2\rightarrow x_3$ are the only edges between nodes $[x_1,x_2,x_3]$, then $x_1$ and $x_2$ cause $x_3$ directly. Therefore, we initialize $C_{3,4}=\{x_1,x_2\}$. There are no possible mediators between $x_3$ and $x_4$, so we initialize $M_{3,4}=\emptyset$. Then, by Lemma \ref{lemma:edge} $x_3\not\ind x_4|Z_{3,4}=x_3\not\ind x_4|x_1,x_2 \iff x_3 \rightarrow x_4$. If an edge exists between $x_3$ and $x_4$, we initialize $M_{2,4}$ accordingly. Note, $x_1$ is a parent of $x_2$, so we initialize $C_{2,4}=x_1$. Then, by Lemma \ref{lemma:edge} $x_2 \not \ind x_4 |Z_{2,4}\iff x_2\rightarrow x_4$. If an edge exists between $x_3$ and $x_4$, and/or between $x_2$ and $x_4$, initialize $M_{1,4}$ accordingly. Note, $x_1$ has no parents, so initialize $C_{1,4}=\emptyset$. Then by Lemma \ref{lemma:edge} $x_1\not\ind x_4|Z_{1,4}\iff x_1\rightarrow x_4$. We recover all possible edges between $[x_1,x_2,x_3,x_4]$.

    \item Finding parents of $x_5$: first we check whether $x_4\rightarrow x_5$, then whether $x_3\rightarrow x_5$, then whether $x_2\rightarrow x_5$, then whether $x_1\rightarrow x_5$.~
    \\
    \\ \vdots
\end{enumerate}

 \paragraph{Iteration $k-1$ Inductive Assumption} We have recovered the edges between $[x_1,\ldots,x_{k-1}]$. 

 \paragraph{Iteration $k$} We now find all nodes in $[x_1,\ldots,x_{k-1}]$ that cause $x_k$ (which yields all edges between $[x_1,\ldots,x_{k}]).$
  \paragraph{Base Case Sub-Iteration (1,2)}
 \begin{enumerate}
     \item  We first check whether $x_{k-1}\rightarrow x_k$. We initialize the potential confounders $C_{k-1,k}$ using$[x_1,\ldots,x_{k-1}]$. The set of mediators $M_{k-1,k}$ is empty by the topological sort. Then, by Lemma \ref{lemma:edge} $x_{k-1}\rightarrow x_{k}$ $\iff$ $x_{k-1}\not\ind x_k | Z_{k-1,k}$.
     \item We now check whether $x_{k-2}\rightarrow x_k$. We initialize the potential confounders $C_{k-1,k}$ using$[x_1,\ldots,x_{k-2}]$. Only $x_{k-1}$ can be a mediator: we know whether $x_{k-2}$ causes $x_{k-1}$, and in our previous step we found whether $x_{k-1}$ causes $x_k$. Thus, we initialize $M_{k-2,k}$ accordingly. Thus, by Lemma \ref{lemma:edge} $x_{k-2}\rightarrow x_{k}$ $\iff$ $x_{k-2}\not\ind x_{k}|Z_{k-2,k}$.
     \\
        \vdots
\end{enumerate}
\paragraph{Sub-Iteration $j$ Inductive Assumption} We have recovered edges between $[x_{j},\ldots,x_k]$
\paragraph{Sub-Iteration $j-1$} We now find if $x_{j-1}$ causes $x_k$.
\begin{enumerate}
\item
For node $x_{j-1}$ where $1\leq j-1 < k$, we obtain $C_{j-1,k}$ by Iteration $k-1$ Inductive Assumption and $M_{j-1,k}$ by Sub-Iteration $j$ Inductive Assumption. Then, by Lemma \ref{lemma:edge} $x_{j-1}\rightarrow x_k \iff x_{j-1}\not \ind x_k | Z_{j-1,k}$.
 \end{enumerate}
Sub-Iteration $j$ Inductive Assumption is satisfied for $j-1$, therefore we recover all nodes in $[x_1,\ldots,x_{k-1}]$ that cause $x_k$. This satisfies Iteration $k-1$ Inductive Assumption for $k$, which means we recover all edges between $[x_1,\ldots,x_k]$. Thus, for a topological sort of arbitrary length, the algorithm recovers all possible edges.

\end{proof}

%We aim to formally extend Edge Discovery to a hierarchical version in later works.

\subsection{Time Complexity for Edge Discovery}\label{appendix:edge_discovery_time}

\edruntime*

\begin{proof}
    ED checks for the existence of every edge permitted by a topological sort $\pi$ by running one conditional independence test that has complexity $O(n^3)$. In the worst case, there are $O(d^2)$ possible edges, so the overall complexity is $O(d^2n^3)$.
\end{proof}

\end{appendices}
\newpage
\begin{appendices}
\section{Additional Experiments and Runtimes}\label{appendix: additional exp}
\subsection{Runtimes for Linear Topological Sort}\label{appendix: lhtsruntime}
~
\begin{figure}[h!]
    \centering
    \begin{subfigure}[b]{0.49\textwidth}
        \includegraphics[width=\textwidth]{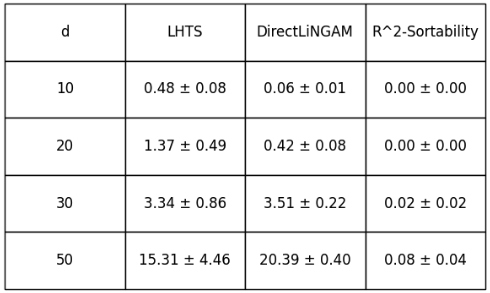}
    \end{subfigure}
     \begin{subfigure}[b]{0.49\textwidth}
        \includegraphics[width=\textwidth]{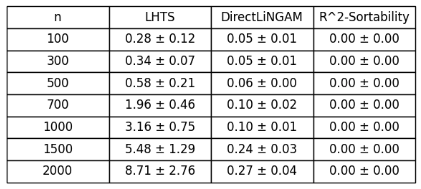}
    \end{subfigure}\caption{Runtimes for linear topological sorts, left: top row; right: bottom row, see Figure \ref{fig: lin_sort}.}\label{fig: lin exp runtimes}
\end{figure}

\subsection{Runtimes for Nonlinear Topological Sort}\label{appendix: nonlin runtimes}
\begin{figure}[h!]
    \centering
        \includegraphics[width=\textwidth]{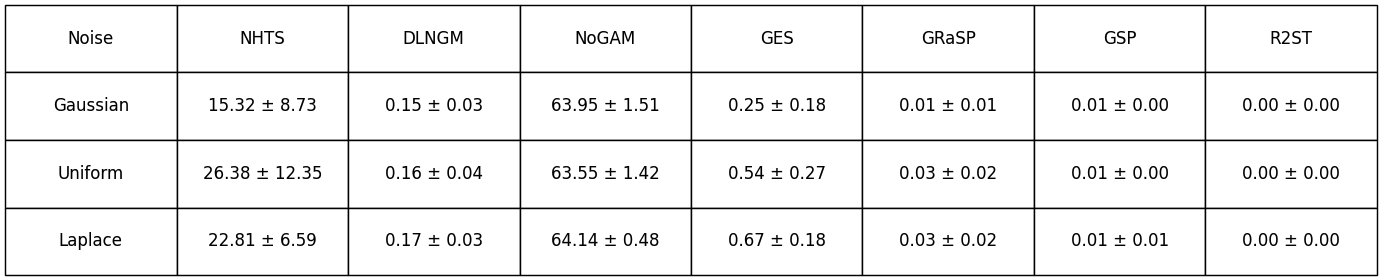}%
        %{figures/nhts_runtimes.png}%{figures/sparseruntimes.png}
  
    \caption{Runtimes for nonlinear topological sorts, see Figure \ref{fig: nonlin_sort}.}\label{fig: nonlin exp runtimes}
\end{figure}

\subsection{Topological Sorts on Nonlinear Data}\label{appendix: nonlinear add exps}

\begin{figure*}[h!]
     \centering
    \begin{subfigure}[t]{0.21\textwidth}  % 1st row, 1st column
        \includegraphics[width=\textwidth]{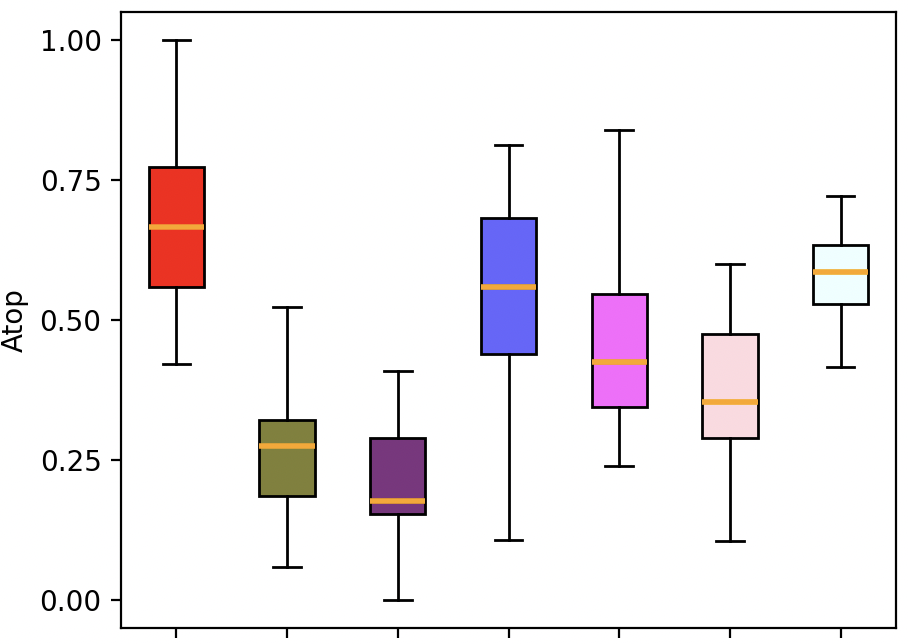}
    \end{subfigure}
    \begin{subfigure}[t]{0.21\textwidth}  % 1st row, 2nd column
        \includegraphics[width=\textwidth]{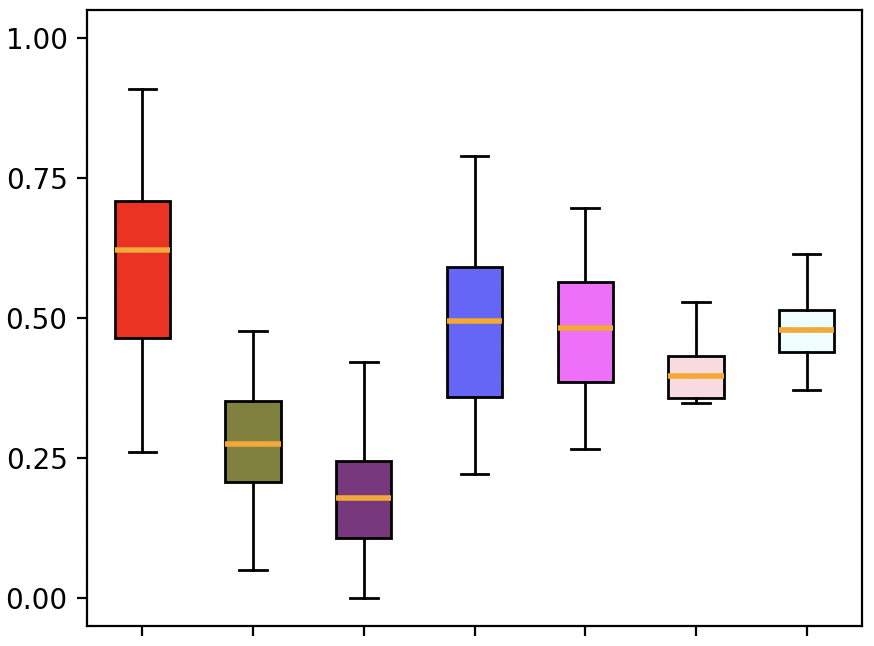}
    \end{subfigure}
    \begin{subfigure}[t]{0.21\textwidth}  % 1st row, 3rd column
        \includegraphics[width=\textwidth]{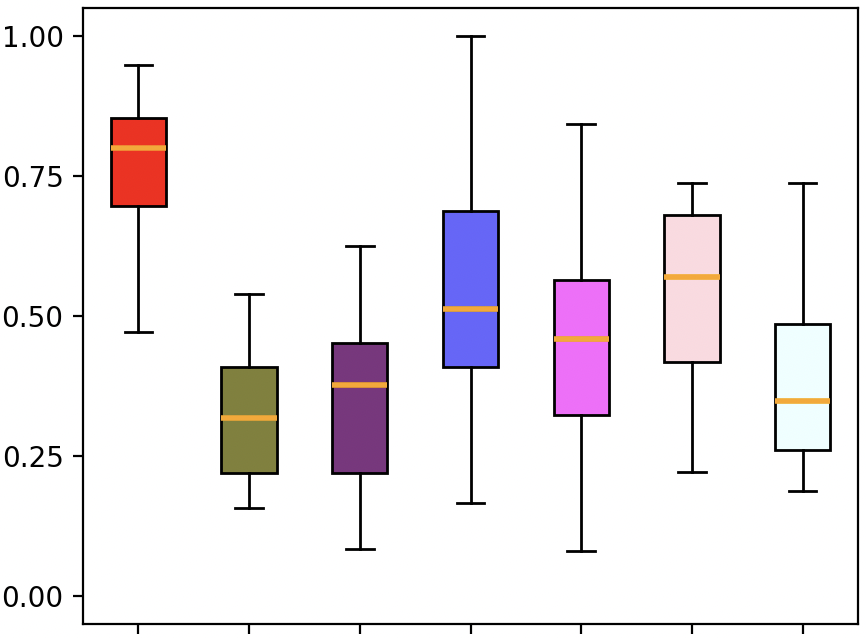}
    \end{subfigure}
    
    \begin{subfigure}[t]{0.21\textwidth}  % 2nd row, 1st column
        \includegraphics[width=\textwidth]{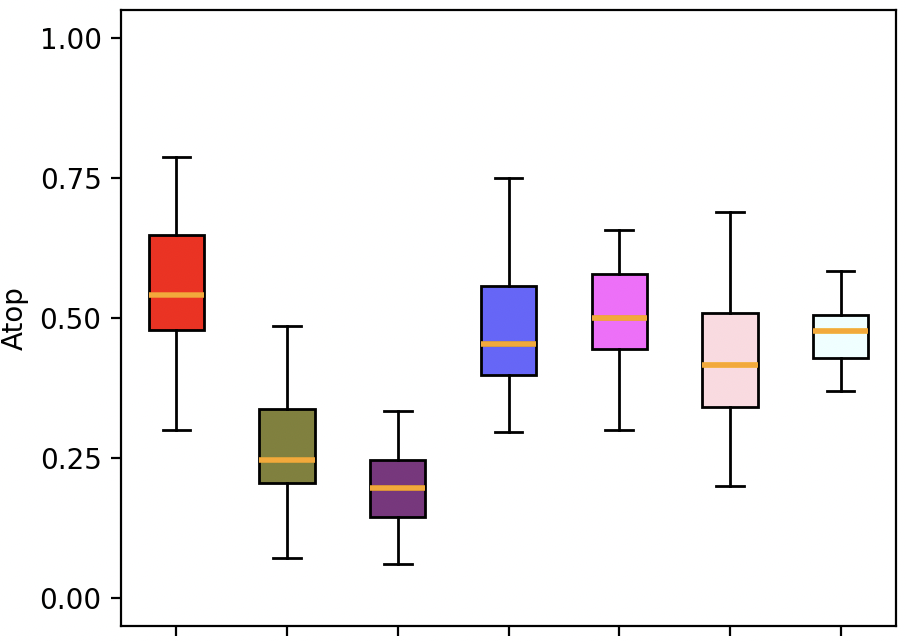}
    \end{subfigure}
    \begin{subfigure}[t]{0.21\textwidth}  % 2nd row, 2nd column
        \includegraphics[width=\textwidth]{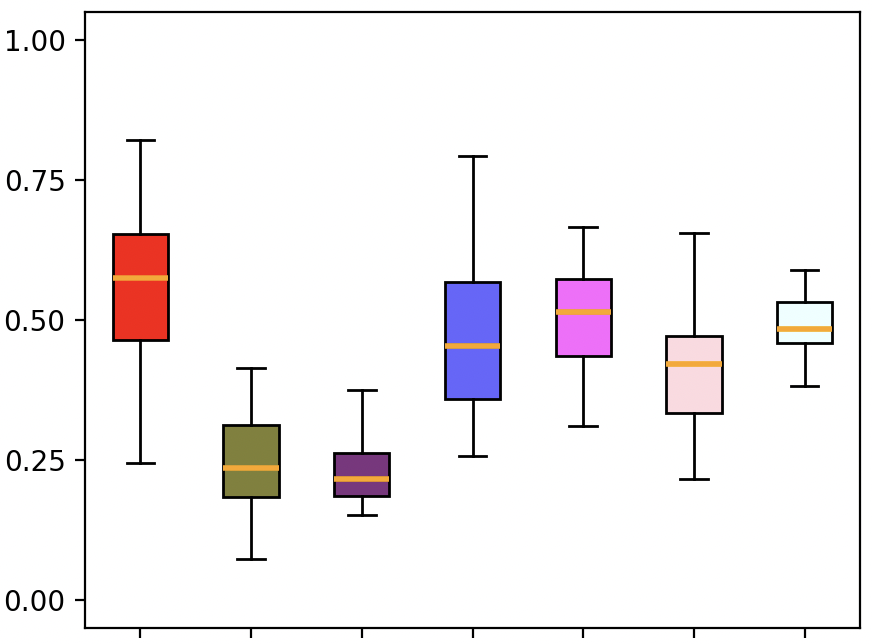}
    \end{subfigure}
    \begin{subfigure}[t]{0.21\textwidth}  % 2nd row, 3rd column
        \includegraphics[width=\textwidth]{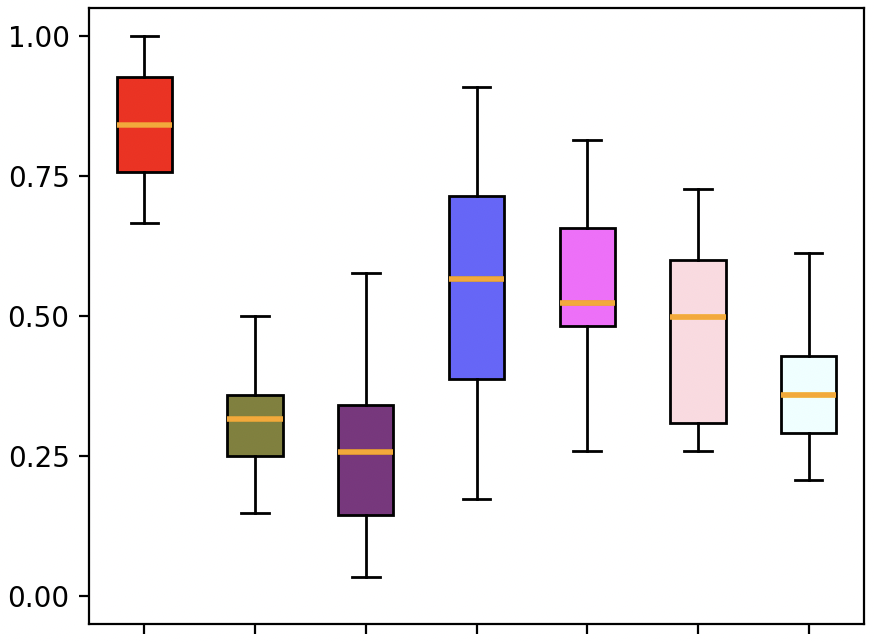}
    \end{subfigure}
    
    \begin{subfigure}[t]{0.21\textwidth}  % 3rd row, 1st column
        \includegraphics[width=\textwidth]{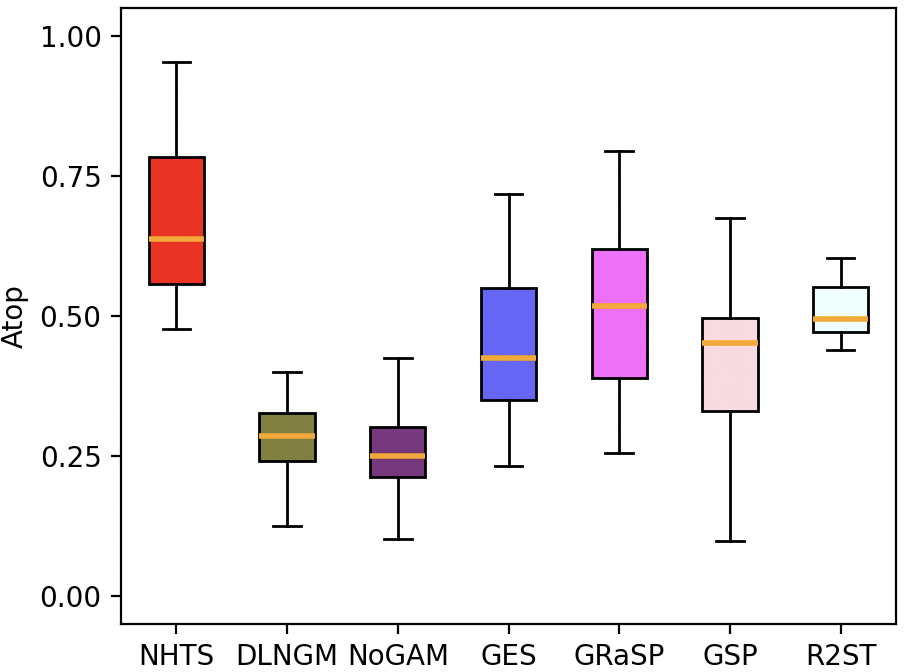}
    \end{subfigure}
    \begin{subfigure}[t]{0.21\textwidth}  % 3rd row, 2nd column
        \includegraphics[width=\textwidth]{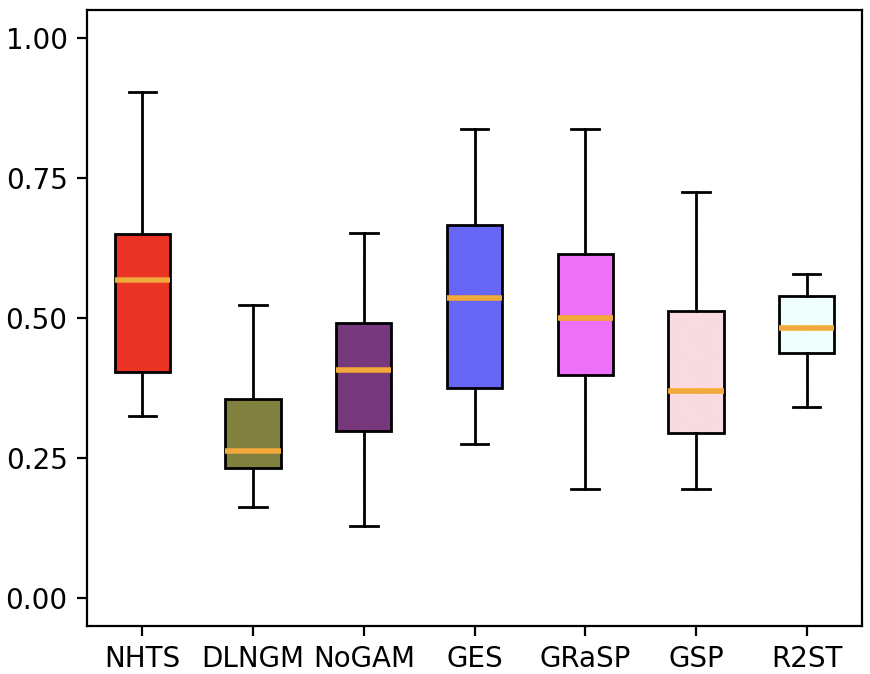}
    \end{subfigure}
    \begin{subfigure}[t]{0.21\textwidth}  % 3rd row, 3rd column
        \includegraphics[width=\textwidth]{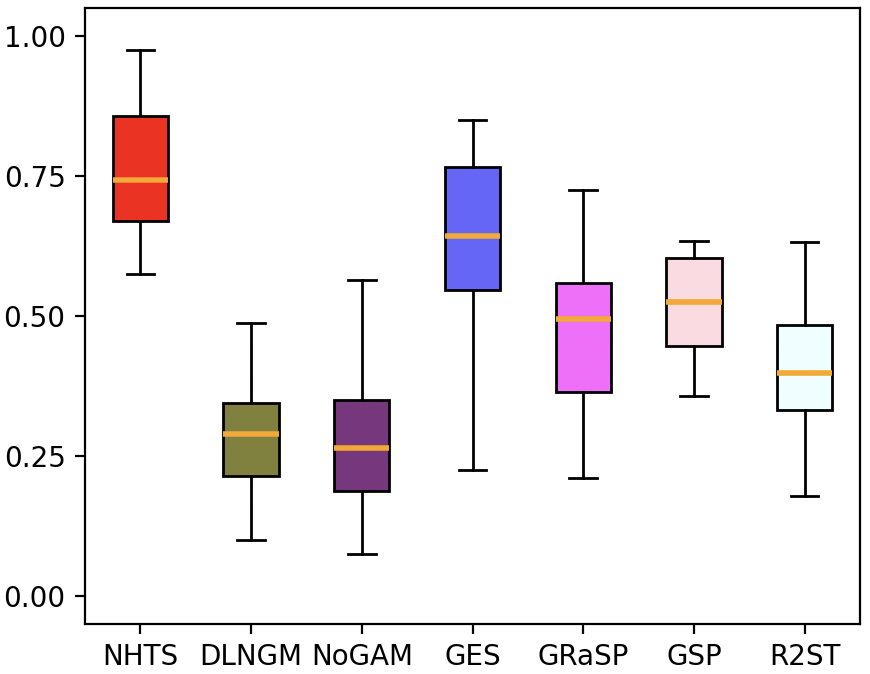}
    \end{subfigure}

    \caption{Performance of NHTS on data generated with Gaussian, Laplace, or Uniform noise (left, middle, right columns), the average number of edges set to $2d$, $3d$, or $4d$ (top, middle, bottom rows).
    }\label{fig: nhts extra exp}
\end{figure*}

In figure \ref{fig: nhts extra exp} we provide additional experiments for nonlinear topological sorts ($d=10,n=300$);
we see that NHTS maintains superior performance even as the density of the underlying graph increases, although the performance gap decreases, especially for laplacian noise, as the graph becomes denser.

\subsection{Edge Pruning on Nonlinear Data}\label{appendix: pruing add exps}

\begin{figure*}[h!]
    \centering
    \begin{subfigure}[t]{0.20\textwidth}  % 1st row, 1st column
        \includegraphics[width=\textwidth]{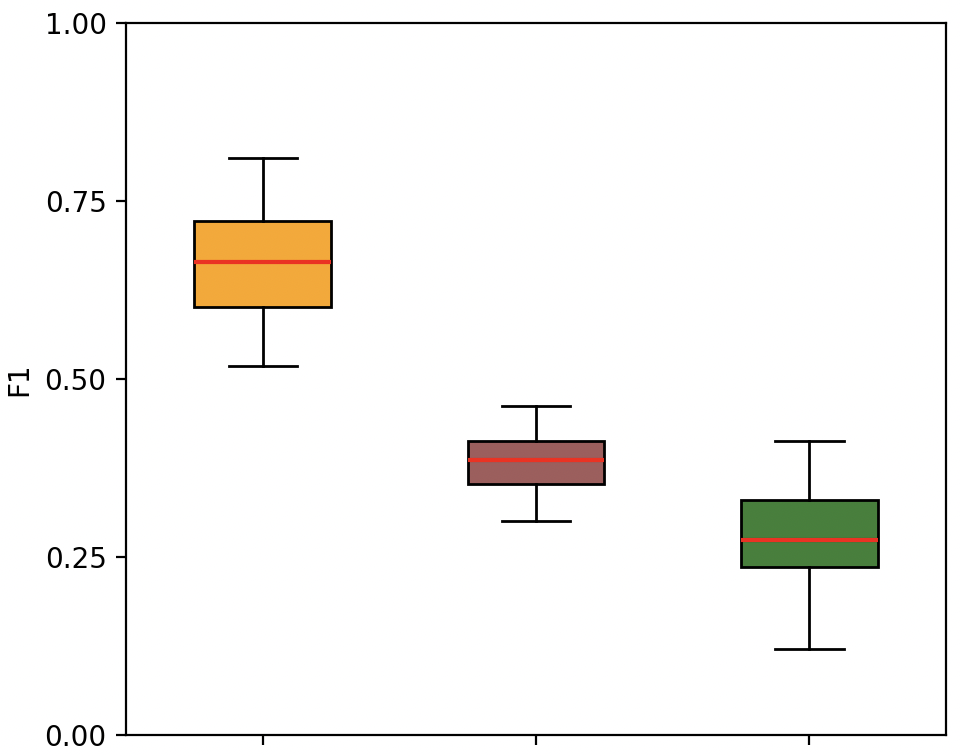}
    \end{subfigure}
    \begin{subfigure}[t]{0.20\textwidth}  % 1st row, 2nd column
        \includegraphics[width=\textwidth]{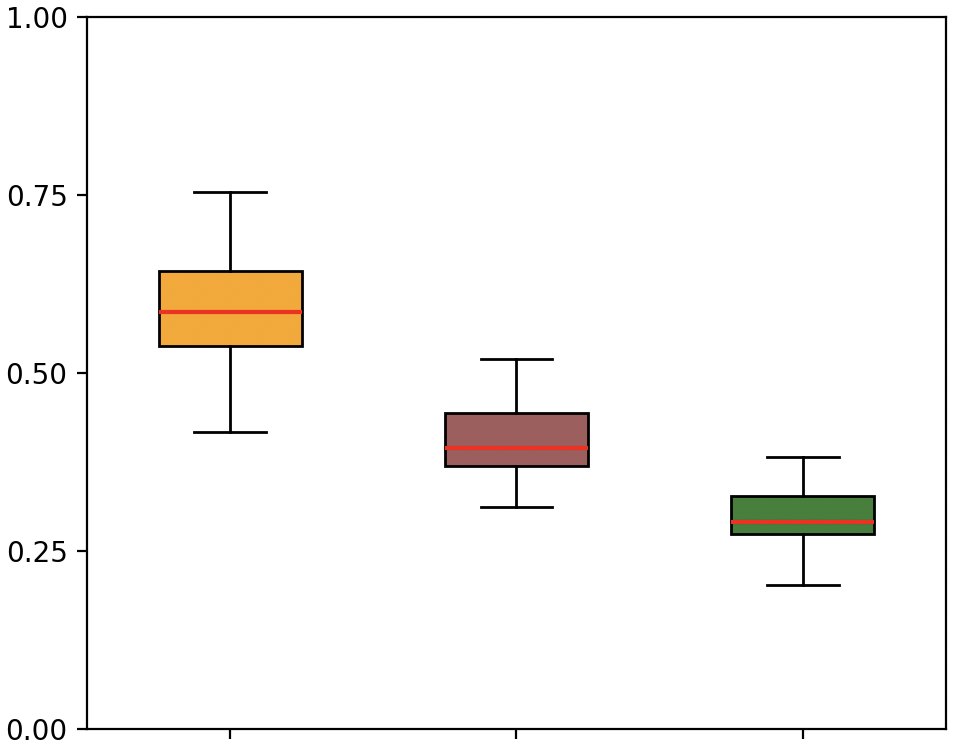}
    \end{subfigure}
    \begin{subfigure}[t]{0.20\textwidth}  % 1st row, 3rd column
        \includegraphics[width=\textwidth]{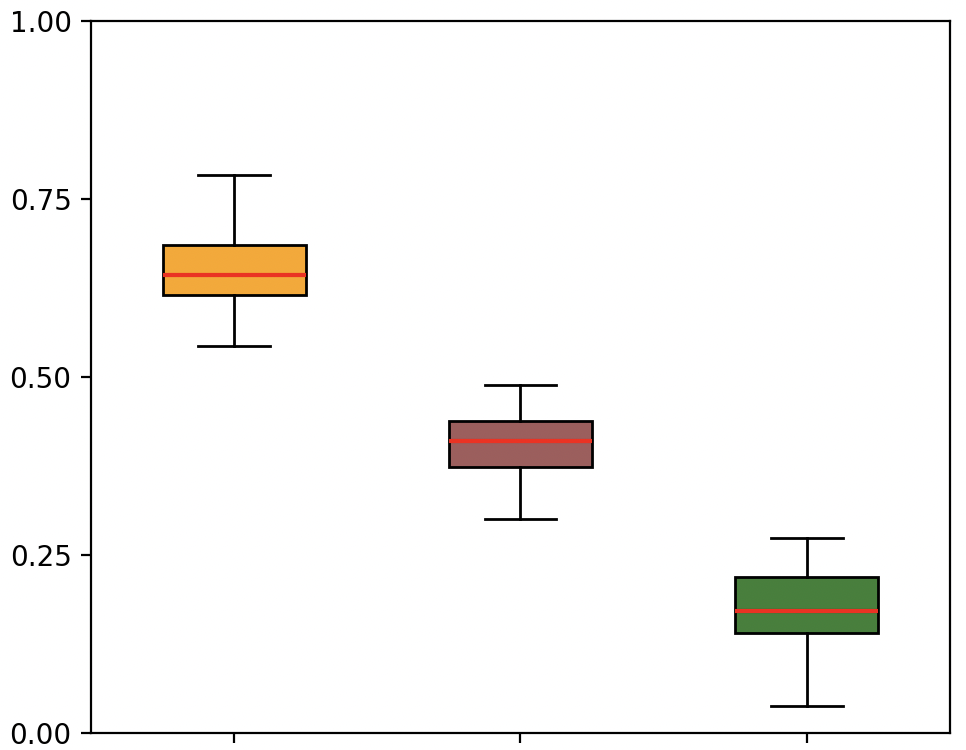}
    \end{subfigure}
    
    \begin{subfigure}[t]{0.20\textwidth}  % 2nd row, 1st column
        \includegraphics[width=\textwidth]{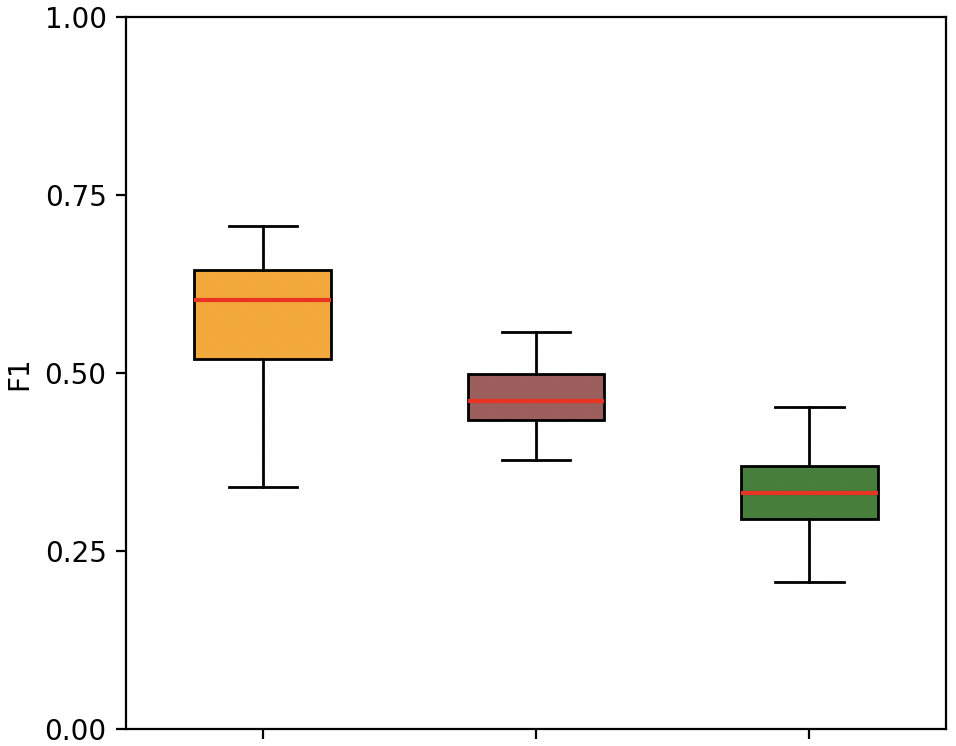}
    \end{subfigure}
    \begin{subfigure}[t]{0.20\textwidth}  % 2nd row, 2nd column
        \includegraphics[width=\textwidth]{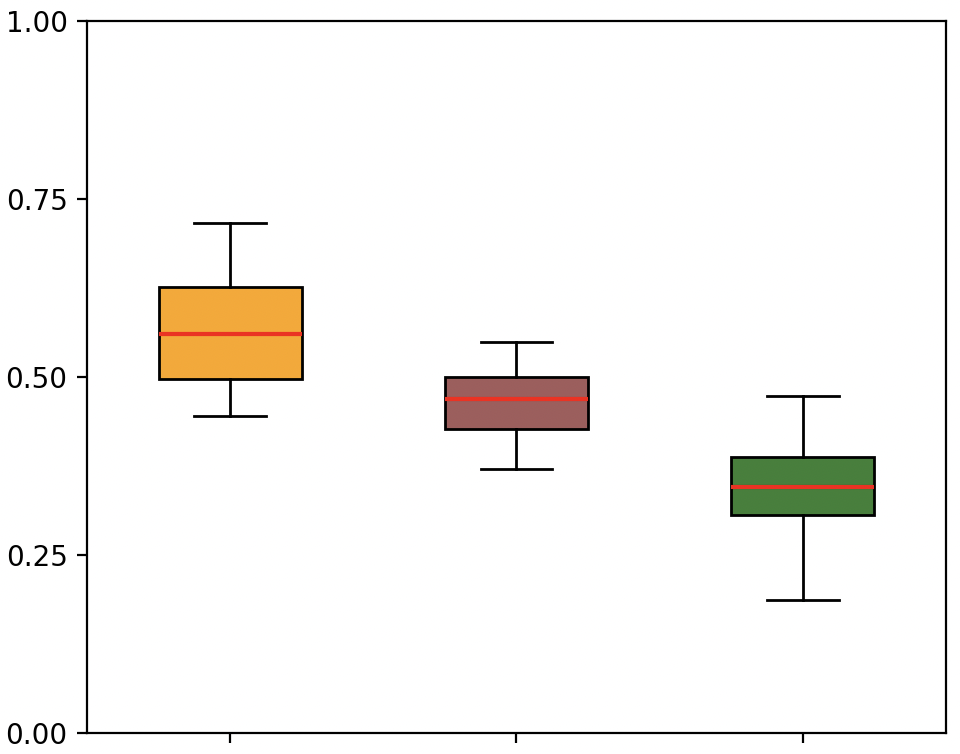}
    \end{subfigure}
    \begin{subfigure}[t]{0.20\textwidth}  % 2nd row, 3rd column
        \includegraphics[width=\textwidth]{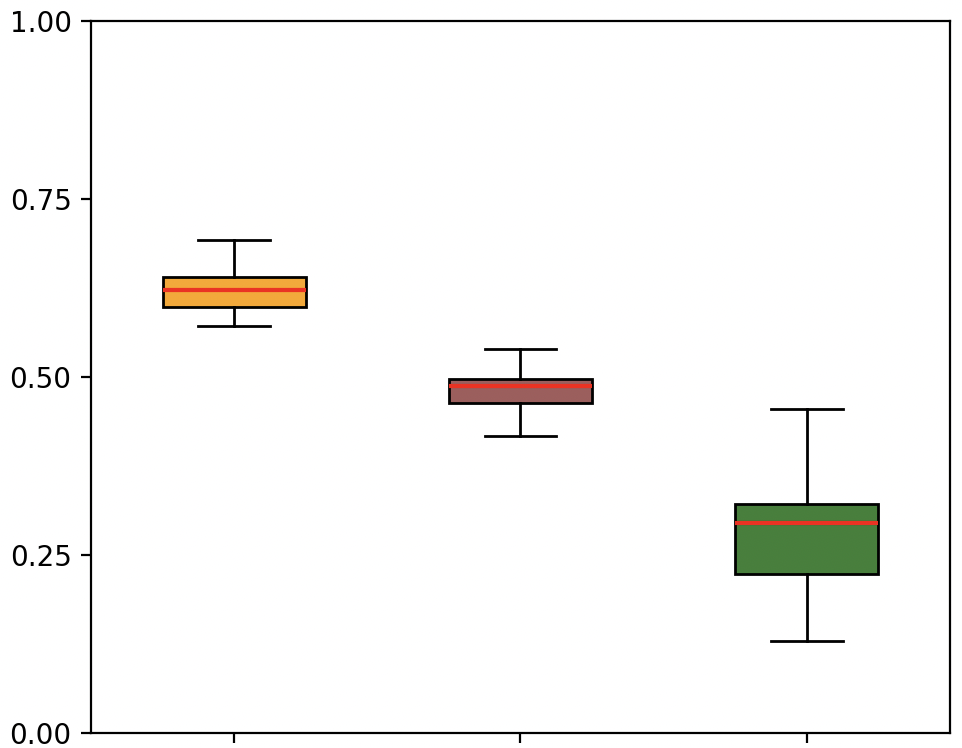}
    \end{subfigure}
    
    \begin{subfigure}[t]{0.20\textwidth}  % 3rd row, 1st column
        \includegraphics[width=\textwidth]{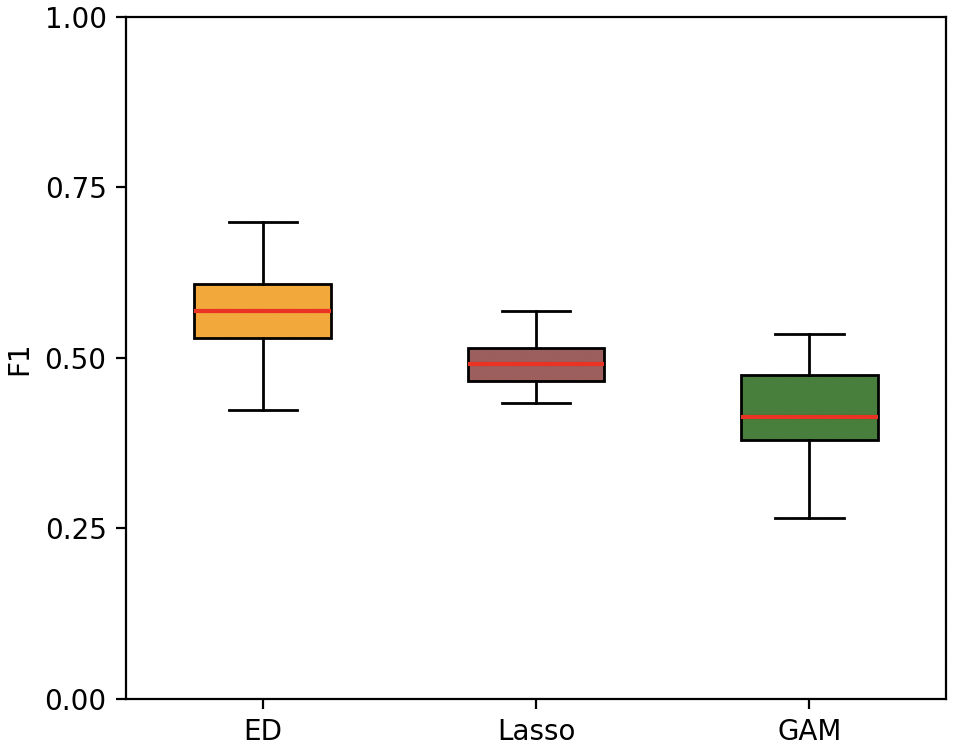}
    \end{subfigure}
    \begin{subfigure}[t]{0.20\textwidth}  % 3rd row, 2nd column
        \includegraphics[width=\textwidth]{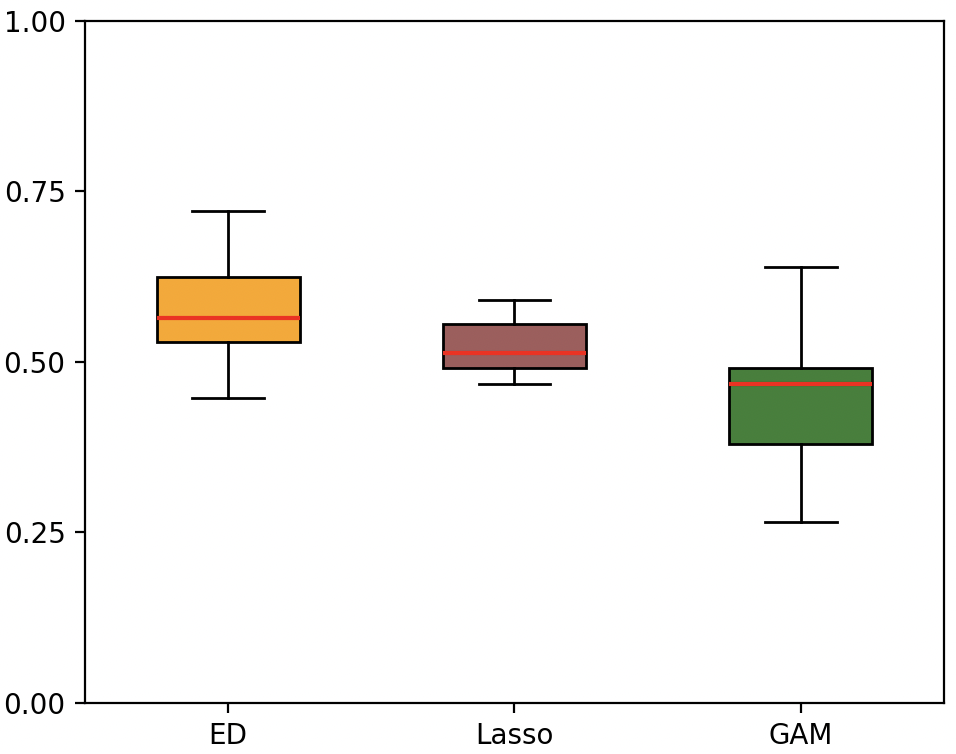}
    \end{subfigure}
    \begin{subfigure}[t]{0.20\textwidth}  % 3rd row, 3rd column
        \includegraphics[width=\textwidth]{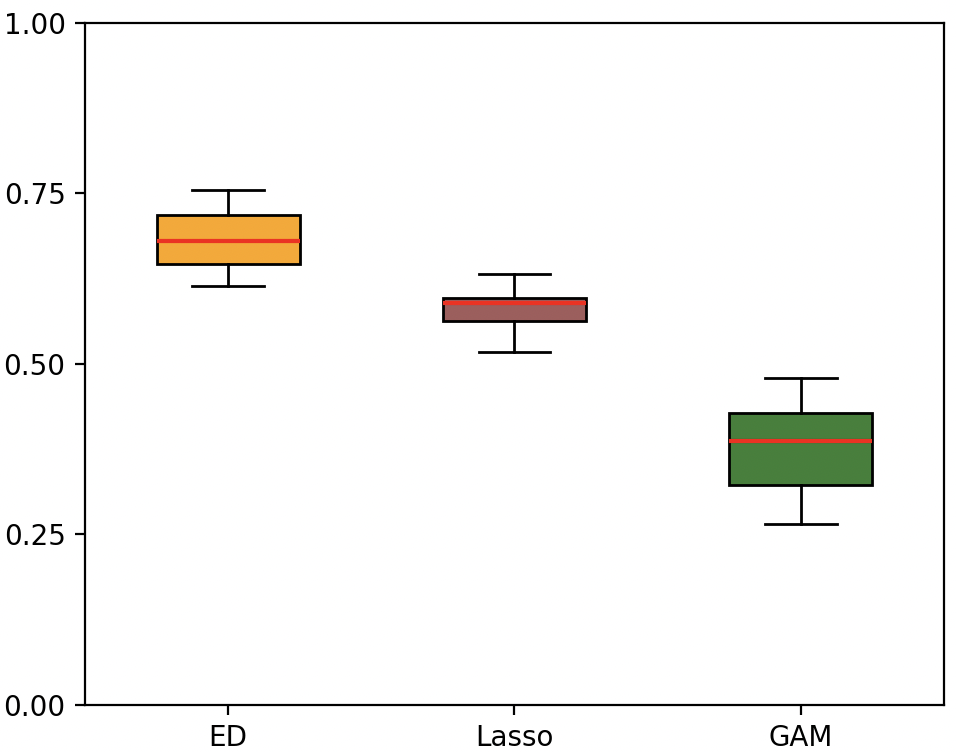}
    \end{subfigure}

    \caption{Performance of ED on data generated with Gaussian, Laplace, or Uniform noise (left, middle, right columns), the average number of edges set to $2d$, $3d$, or $4d$ (top, middle, bottom rows).
    } \label{fig: ED extra exp}
\end{figure*}

In figure \ref{fig: ED extra exp} we provide additional experiments for edge pruning methods ($d=20,n=300$); ED generally maintains superior performance as the noise distribution is varied and density is increased, although the performance gap decreases for all noise distributions as the graph becomes denser.

%The runtime chart on the left corresponds to the top row of graphs in Figure \ref{fig: lin_sort}; The runtime chart on the right corresponds to the bottom row of graphs on the right in Figure \ref{fig: lin_sort}.

\subsection{Runtimes for Edge Pruning}\label{appendix: edgepruneruntimes}
\begin{figure}[h!]
    \centering
    \begin{subfigure}[b]{0.32\textwidth}
        \includegraphics[width=\textwidth]%{figures/resitgeddisc44.png}
        {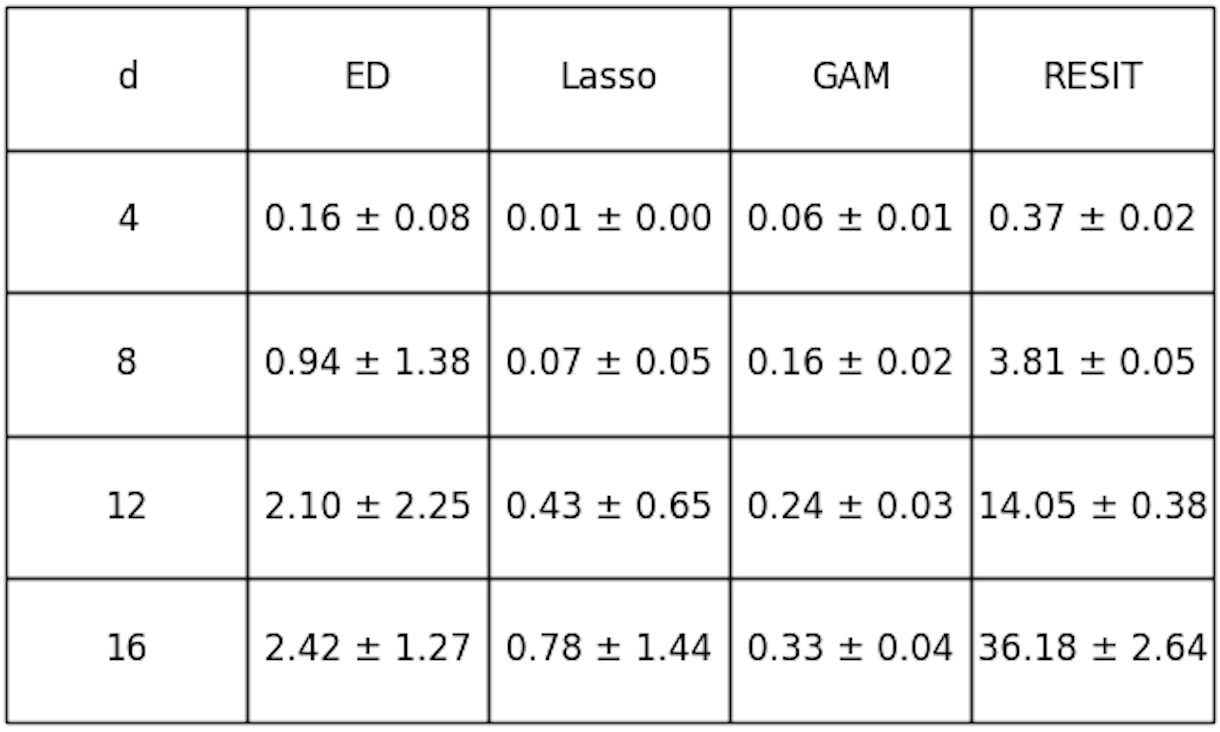}
    \end{subfigure}
    \begin{subfigure}[b]{0.32\textwidth}
        \includegraphics[width=\textwidth]%{figures/holyed44.png}
    {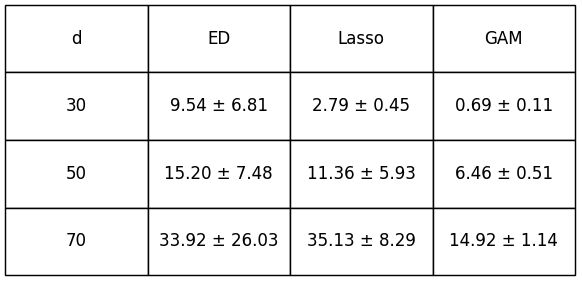}
    \end{subfigure}
    \begin{subfigure}[b]{0.32\textwidth}
        \includegraphics[width=\textwidth]%{figures/edgerprun3runtimee.png}
        {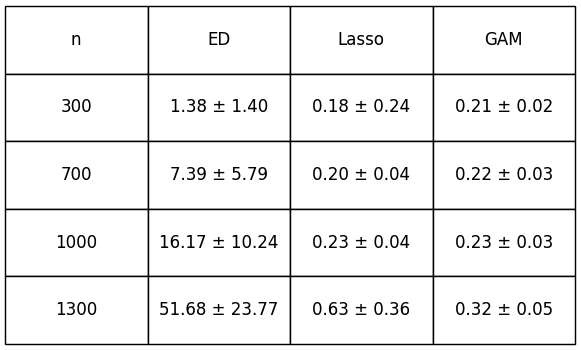}
    \end{subfigure}
    \caption{Runtimes for edge pruning: tables correspond to the graphs in Figure \ref{fig: edge_prune}, from left to right.}\label{fig: ed exp runtimes} 
\end{figure}

\end{appendices}
\newpage
\begin{appendices}
\section{Implementation Details}\label{appendix:experiment implement}
All tests were done in Python. All runtimes were computed locally on an Apple M2 Pro Chip, 16 Gb of RAM, with no parallelization.

%Runtimes in Figure \ref{fig: lin exp runtimes} and Figure \ref{fig: ed exp runtimes} were computed locally on an Apple M2 Pro Chip, 16 Gb of RAM, with no parallelization. Runtimes in Figure \ref{fig: nonlin exp runtimes} were computed on an AWS EC2 instance of type c5.24xlarge.

\subsection{Topological Sort for LiNGAM}\label{appendix:lingam_exp}
\raggedright
\begin{comment}
\begin{enumerate}[leftmargin=*, itemsep=0pt, parsep=0pt]
    \item DirectLiNGAM was imported from the lingam package (https://lingam.readthedocs.io/en/latest/reference/index.html).
    \item  $R^2-$sort was imported from the CausalDisco package (https://github.com/CausalDisco/CausalDisco).
    \item  LHTS was implemented using the LinearRegression function using the Sklearn linear models package (https://scikit-learn.org/stable/modules/linear\_model.html)
    and used independence tests from the PyRKHSstats package, with cutoffs $\alpha = 0.1$.
\end{enumerate}
\end{comment}
 %See file named 'topological\_sort\_experiments' for further details. 
 DirectLiNGAM was imported from the \texttt{lingam} \citep{lingam_package} package, $R^2-$sort was imported from the \texttt{CausalDisco} \citep{reisach_beware_2021, reisach_scale-invariant_2023} package, and LHTS was implemented using the \texttt{Sklearn} \citep{scikit-learn} package. All assets used have a CC-BY 4.0 license. We follow \citep{shimizu2011directlingam} to generate the data for Figure \ref{fig: lin_sort}, using linear causal mechanisms with randomly drawn coefficient values, plus independent uniform noise. Data is standardized to remove shortcuts \citep{reisach_beware_2021}. See Github repository for more details. Cutoff values for independence tests were set to $\alpha = 0.05$ for all methods.
\subsection{Topological Sort for Nonlinear ANM}\label{appendix:nhts_exp}
\raggedright
 GES and GRaSP were imported from the \texttt{causal-learn} \citep{zheng2024causal-package} package; GSP was imported from the \texttt{graphical\_model\_learning} \citep{gsp_package} package. DirectLiNGAM was imported from the \texttt{lingam} package \citep{lingam_package}. NoGAM was imported from the \texttt{dodiscover} \citep{Li_Dodiscover_Causal_discovery} package. $R^2-$sort was imported from the \texttt{CausalDisco} package. NHTS and LoSAM were implemented using the kernel ridge regression (KRR) function from the \texttt{Sklearn} package, used independence tests from either the \texttt{causal-learn} package or the \texttt{dcor} \citep{dcor-ind-package} package, and a mutual information estimator from the \texttt{npeet} \citep{npeet_package} package. All assets used have a CC-BY 4.0 license. We follow \citep{maasch2024local} to generate the data used for Figure \ref{fig: nonlin_sort} and Figure \ref{fig: nhts extra exp}, using quadratic causal mechanisms with randomly drawn coefficient values, plus independent gaussian, laplace or uniform noise. Features generated with quadratic mechanisms were standardized after being generated to remove shortcuts \citep{reisach_beware_2021} and to prevent the quadratic mechanisms from driving all values close to 0 (ensuring stability). See Github repository for more details. Note that, to enable a fair comparison between NHTS and other topological ordering methods, we implement a version of NHTS that returns a linear topological sort, rather than a hierarchical topological sort, by adding only one vertex to the sort in each iteration of its sorting procedure. For NHTS, in Stage 2 we used KRR
 with polynomial kernel, $\alpha = 1$, degree = $3$, coef0 = $1$, and in Stage 4 we used KRR with RBF kernel, $\alpha = 0.1$, $\gamma = 0.01$.
 Cutoff values for independence tests were set to $\alpha = 0.05$ for all methods, no cross validation was allowed for any method. Otherwise, default settings were used for all baselines.
\begin{comment}

\begin{figure}[h!]
    \centering
        \includegraphics[width=\textwidth]{figures/sparseruntimes.png}%{figures/nhts_runtimes.png}
  
    \caption{Runtimes for nonlinear topological sorts, see Figure \ref{fig: nonlin_sort}.} 
\end{figure}
\end{comment}
\subsection{Edge Pruning}\label{appendix:edge_exp}
\raggedright
\begin{comment}
\begin{enumerate}[leftmargin=*, itemsep=0pt, parsep=0pt]
    \item Lasso Regression was implemented using the Sklearn linear models package (https://scikit-learn.org/stable/modules/linear\_model.html).
    \item Hypothesis testing with GAMs was implemented using Bsplines and GLMGam from the Statsmodels package (https://www.statsmodels.org/stable/index.html).
    \item  RESIT was implemented using the RandomForestRegressor function using the Sklearn ensemble package (https://scikit-learn.org/stable/modules/ensemble.html) as indicated by https://lingam.readthedocs.io/en/latest/tutorial/resit.html\#model.
    \item Either the Dcor package (https://dcor.readthedocs.io/en/latest/) or the causallearn CIT package (https://causal-learn.readthedocs.io/en/latest/) was used for independence tests.
\end{enumerate}
\end{comment}
Lasso and RESIT were implemented using the \texttt{sklearn} package, hypothesis testing with GAMs was implemented using Bsplines and GLMGam from the \texttt{statsmodel} \citep{statsmodels-package} package. Independence tests used either the \texttt{causal-learn} package or the \texttt{dcor} package.
All assets used have a CC-BY 4.0 license. We follow \citep{maasch2024local} to generate the data used for Figure \ref{fig: edge_prune} and Figure \ref{fig: ED extra exp}, using quadratic causal mechanisms with randomly drawn coefficient values, plus independent uniform, gaussian, or laplace noise. Features generated with quadratic mechanisms were standardized after being generated to remove shortcuts \citep{reisach_beware_2021} and to prevent the quadratic mechanisms from driving all values close to 0 (ensuring stability). See Github repository for more details. Cutoff values for independence tests were set to $\alpha = 0.05$ for all methods.

\begin{comment}

\begin{figure}[h!]
    \centering
    \begin{subfigure}[b]{0.32\textwidth}
        \includegraphics[width=\textwidth]{figures/resitedgedisc4.png}
    \end{subfigure}
    \begin{subfigure}[b]{0.32\textwidth}
        \includegraphics[width=\textwidth]{figures/holyed4.png}
    \end{subfigure}
    \begin{subfigure}[b]{0.32\textwidth}
        \includegraphics[width=\textwidth]{figures/edgerprun3runtime.png}
    \end{subfigure}
    \caption{Runtimes for edge pruning: tables correspond to the graphs in Figure \ref{fig: edge_prune}, from left to right.} 
\end{figure}
\end{comment}
\end{appendices}

\end{document}